\documentclass[12pt,a4]{article}
\usepackage{amsmath}
\usepackage{amssymb}
\usepackage{amsfonts}
\usepackage{array}
\usepackage{graphicx}
\usepackage{mathrsfs}
\usepackage{multirow}
\usepackage{siunitx}
\usepackage{subcaption}
\usepackage{array}
\newcommand{\bea}{\begin{eqnarray*}}
\newcommand{\eea}{\end{eqnarray*}}
\newcommand{\beao}{\begin{eqnarray}}
\newcommand{\eeao}{\end{eqnarray}}

\usepackage{biblatex}
\title{Applications and Manipulations of Physics-Informed Neural Networks in Solving Differential Equations}

\author{
Aarush Gupta, Kendric Hsu, Syna Mathod}

\begin{document}
\maketitle

\begin{abstract}
Mathematical models in neural networks are powerful tools for solving complex differential equations and optimizing their parameters; that is, solving the forward and inverse problems, respectively. A forward problem predicts the output of a network for a given input by optimizing weights and biases. An inverse problem finds equation parameters or coefficients that effectively model the data. A Physics-Informed Neural Network (PINN) can solve both problems. PINNs inject prior analytical information about the data into the cost function to improve model performance outside the training set boundaries. This also allows PINNs to efficiently solve problems with sparse data without overfitting by extrapolating the model to fit larger trends in the data. The prior information we implement is in the form of differential equations. Residuals are the differences between the left-hand and right-hand sides of corresponding differential equations; PINNs minimize these residuals to effectively solve the differential equation and take advantage of prior knowledge. In this way, the solution and parameters are embedded into the loss function and optimized, allowing both the weights of the neural network and the model parameters to be found simultaneously~—~solving both the forward and inverse problems in the process. In this paper, we will create PINNs with residuals of varying complexity, beginning with linear and quadratic models and then expanding to fit models for the heat equation and other complex differential equations. We will mainly use Python as the computing language, using the PyTorch library to aid us in our research.
\end{abstract}

\section{Introduction}
In recent years, the field of deep learning has advanced immensely, especially by applying neural networks to complex problems that require prior information to solve~\cite{janiesch_machine_2021, schmidhuber_deep_2015}. Deep learning, a specific type of machine learning, employs neural networks with hidden layers capable of using advanced convolutions in each neuron to automatically find representations from raw, high-dimensional data~\cite{janiesch_machine_2021}. Scientific Machine learning (SciML) focuses on integrating specific knowledge and scientific principles about the problem at hand with machine learning techniques~\cite{cuomo_scientific_2022}. A prominent use case of SciML is Physics-Informed Machine Learning (PIML), which refers to incorporating physical laws or constraints into machine learning models~\cite{cuomo_scientific_2022}. This ensures that models are consistent and suitable for applications while adhering to physical principles~\cite{cuomo_scientific_2022, wolf_physics-informed_2024}.

A specific example of PIML is the Physics-Informed Neural Network (PINN), which is a novel class of neural networks that integrates known physical laws and equations in the form of differential equations~\cite{wolf_physics-informed_2024, ben_moseley_article}. 
PINNs operate by embedding differential equations into the loss function. In this way, the network both fits the observed data and adheres to the underlying physical laws governing the system~\cite{ben_moseley_article}. In \cite{ben_moseley_article}, the model makes accurate, reliable predictions; it can maintain high levels of performance even with insufficient and/or costly data.

PINNs are versatile and can be applied to various scientific and engineering concepts, based on mathematical models. For instance, PINNs have been used to solve complex and relevant problems such as modeling earthquake waves, fluid dynamics, and molecular systems.~\cite{ben_moseley_video} In the comprehensive review of PINNs by Cuomo et al.~\cite{cuomo_scientific_2022}, the authors describe recent advancements in PINNs, including customizing network architectures, optimization techniques, and loss function structures.

A residual is the difference between the left and right-hand sides of a differential equation. The residual is useful because when it is zero, we satisfy the differential equation. By adjusting parameters such as the number of layers, neurons, or the strength of the residual in our neural networks, we can better tailor the network toward specific applications and problems to solve. In this paper, we detail our experimentation with PINNs and examine the effects of residual strength on our ability to fit data. The strength of the residual in the loss function influences the importance of our prior information. The exact residual of the differential equation changes based on the type of problem we attempt to solve. 

Ultimately, we aim to develop a PINN that can accurately estimate parameters for the Heat Differential Equation and eventually the GBM Partial Differential Equation~\cite{zhang_personalized_2024} with limited training data points. This approach will illustrate the potential of PINNs in solving complex differential equations, such as predicting the spatial dynamics of GBM tumors for individualized medical treatments and science elsewhere.


\subsection{Differential Equations}

In this paper, we fit various PINNs to datasets consistent with some differential equations. However, all differential equations that describe the data fall under the general notation:
\begin{equation}\label{eq:generalde}
    P(\partial_t, \partial_x, ... ; u)u = 0
\end{equation}
Equation \ref{eq:generalde} describes a differential equation as a product between two functions $P$ and $u$. $P$ is a function of relevant derivative operators with respect to $t$ and $x$. $u$ is a function of one or more variables. Equation \ref{eq:generalde}'s notation is flexible and can represent both ordinary differential equations (ODEs) with one variable and partial differential equations (PDEs) with multiple variables. 

Not all differential equations we encounter here are linear, so at times, $u$ may also be a component of $P$. A linear equation indicates that the highest degree of $u$ in the function is $1$, but we will see higher degrees later on.

The residual of a differential equation is essential in implementing prior knowledge into a PINN. The residual of an equation following Equation \ref{eq:generalde}'s structure is the difference between the left-hand side and the right-hand side, which is $P(\partial_t, \partial_x, ... ; u)u$. We can solve a differential equation by adding such a residual to the loss function and then minimizing it algorithmically.

\subsubsection{Linear Equation}\label{sec:linear}
If we believe the data has a linear relationship, we rely on the following equations in our PINN:
\begin{equation}\label{eqn:linear}
m(x) = ax + b
\end{equation}
\begin{equation}\label{eqn:linear_diffeq}
\frac{d^2m}{dx^2} = 0
\end{equation}

Equation \ref{eqn:linear} describes a basic linear function, where \(a\) and \(b\) are any real numbers. Equation \ref{eqn:linear_diffeq} represents the second derivative of the linear function, which is zero due to its linear nature. 

\subsubsection{Quadratic Equation}
If we believe the data has a quadratic relationship, we rely on these equations in our PINN:
\begin{equation}\label{eqn:quad}
n(x) = ax^2 + bx + c
\end{equation}
\begin{equation}\label{eqn:quad_diffeq}
\frac{d^3n}{dx^3} = 0
\end{equation}

Equation \ref{eqn:quad} describes a parabolic function where \(a\), \(b\), and \(c\) are real numbers. Equation \ref{eqn:quad_diffeq} is its third derivative, which equals zero because of the quadratic nature of the function. 

But we are not just limited to first or second degree polynomials. For an \( n \)-th degree polynomial \( f(x) \):

\[ f(x) = a_n x^n + a_{n-1} x^{n-1} + \cdots + a_1 x + a_0 \]

\[ f^{(n+1)}(x) = 0 \]

We can see that the \((n+1)\)-th derivative of \( f(x) \) will be zero:

This is because each derivative reduces the degree of the polynomial by 1. Therefore, for a polynomial of degree \( n \), the \((n+1)\)-th derivative is zero.

This property is useful in PINNs, where we leverage the fact that certain derivatives of our functions will be zero. We can then incorporate this information into the loss function of the neural network. This helps enforce the underlying data trends described by differential equations.

\subsubsection{Heat Equation}\label{sec:heat}
If we have data that describes temperature in one dimension space with respect to time, we consider this one-dimensional heat equation in our PINN:
\begin{equation}\label{eq:heat}
    \frac{\partial u}{\partial t} = D \frac{\partial^2 u}{\partial x^2}
\end{equation}
As described in~\cite{noauthor_heat_nodate}, Equation \ref{eq:heat} models the temperature \(u(x, t)\) of a one-dimensional rod with length from \(x = 0\) to \(x = L\), where \(D > 0\) is the thermal diffusivity coefficient.


\section{Methods}

\subsection{Data}
Suppose we have some number of data points that are consistent with some differential equation. We generate the data differently based on the problem we are solving. For the linear and quadratic models, we generate synthetic data by adding Gaussian noise to a specified curve like Equations \ref{eqn:linear} and \ref{eqn:quad}. We use a similar process to create synthetic data when modeling the heat equation. 

\subsection{Neural Network}\label{subsec:nn}
We begin by building a PINN that aims to fit the data with a solution that is consistent with the differential equation. We define the network with an optimal number of layers and neurons for the problem at hand. To determine this configuration, we begin with a certain number of layers and then change the number of neurons. After finding the number of neurons with the smallest loss, we manipulate the number of layers. The final result should yield a configuration that fits our data accurately. 

For the linear and quadratic models, we start with 4 total layers and then try models with 5, 10, 20, 30, and 40 neurons per layer. After finding a satisfactory number of neurons that results in low loss, we try 2, 3, 4, 5, and 6 total layers. The number of layers and neurons with the best performance serves as our configuration. 

For the heat equation model, we start with 4 total layers but try models with 10, 20, 40, 60, and 80 neurons per layer. We increase the neuron counts in the models because we anticipate the higher complexity of the heat equation data. More neurons can allow the model to learn more complex data. In our process, we then move on with a satisfactory number of neurons and again test 2, 3, 4, 5, and 6 total layers.

We next train the model over 5000 epochs using the Adam optimizer. Adam uses momentum to accelerate gradient descent by keeping track of an exponentially decaying average of past gradients~\cite{vishwakarma_what_2023}. Using the momentum it gains, Adam becomes very efficient in finding the minima of the cost surface~\cite{vishwakarma_what_2023, kingma_adam_2017}. Every epoch, we pass the data through the model and calculate its loss with the following equation:

\begin{equation}
    L_{data}= \frac{1}{n} \sum_{i=1}^{n}(Y_{i}-\hat{Y}_{i})^2
\end{equation}

We use automatic differentiation to compute the loss or residual of the differential equation. Below is the general form:

\begin{equation}
    {L_{residual} = P(\partial_t, \partial_x, ... ; u)u}
\end{equation}

We add the data loss and residual loss together, and in the overall loss function, we set the strength of the residual $\lambda$ to control how much the model is influenced by prior information. 

\begin{equation}
    L_{total} = L_{data} + \lambda(L_{residual})
\end{equation}

We differentiate the loss with respect to the weights and biases and use those derivatives to modify the model parameter values and improve performance. After training, the loss should have decreased significantly.

\begin{figure}[htp]
    \centering
    \makebox[\textwidth][c]{\includegraphics[width=1.0\textwidth]{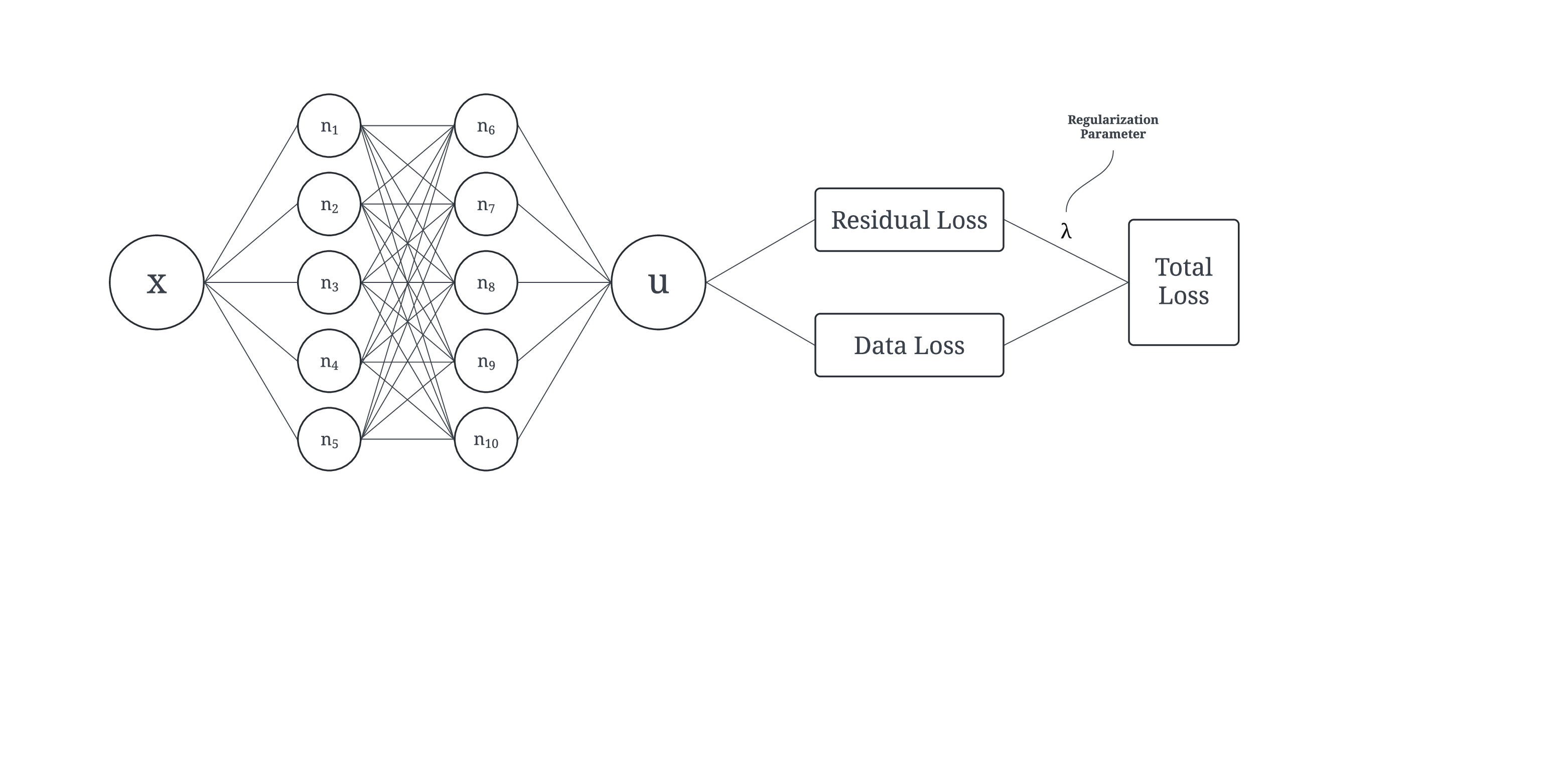}}
    \caption{Schematic of PINN. $\lambda$ is the regularization parameter. PINNs are used to solve differential equations and estimate parameters from data. $n_{1 - 10}$ represent the neurons in the PINN. $x$ and $u$ are the inputs and outputs, respectively.}
    \label{fig:PINN_Schematic}
\end{figure}

However, there are a few limitations to using this method of finding optimal neural network architecture. To begin with, we are not able to test and record the losses of every possible combination. A model with a different combination of neurons/layers may potentially be more optimal. Another limitation of this method is that it relies on a ground truth function to calculate error. If we do not know this function, we would use the model's total loss which may result in a different optimal model. With all this in mind, it must be noted that a model that fits data without significant computational expense is sufficient for our purposes.

\subsection{Alternative Approach to PINNs}\label{subsec:fdm}
When working with the heat equation, we will explore another approach to solving PDEs called a finite difference method (FDM). This method involves updating the solution for a given index based on the previous index's solution. A discrete formula for the solution function is obtained by substituting finite differences for derivatives. The final formula is below:
\begin{equation}
    u_i^{n+1} = u_i^n + \frac{D \Delta t}{(\Delta x)^2} \left( u_{i+1}^n - 2u_i^n + u_{i-1}^n \right)
\end{equation}
An important aspect of an FDM is that a Courant-Friedrichs-Lewy (CFL) condition must be satisfied to produce stable and accurate results. This condition for the heat equation enforces the shape of the ground truth function and the density of the data points:
\begin{equation}
    D \Delta t \le 0.5(\Delta x)^2
\end{equation}
We implement an FDM in MATLAB in addition to our PINN. Considering this CFL condition, we  test different values of $\Delta x$, $\Delta t$, $D$ to investigate where the FDM and PINN stop producing reasonable results. We expect that the ground truth MSE of the FDM will blow up past the CFL condition, but we also look at the ground truth MSE of the PINN to see if it follows the same trend. In addition to quantitative analysis of the loss, we also look at 3D plots of the solution surfaces to define a threshold.

To begin, we define three main types of models that apply to both the FDM and PINN: one with $\Delta x = \Delta t = 0.1$, another with $\Delta x = \Delta t = 0.05$, and a final one with $\Delta x = \Delta t = 0.025$. Within those densities, we vary the final time that the model runs to, testing $t=1$, $t=10$, and $t=100$. Within those times, we test at least three D values: one that satisfies the condition, another that satisfies it at the boundary, and a final one that breaks it. Our goal is to explore the relationship between an FDM and a PINN.

\section{Results}
\subsection{Linear PINN}
We set up our first PINN to learn from noisy synthetic data that is consistent with the differential equation belonging to a line: \begin{equation}
\frac{d^2m}{dx^2} = 0
\end{equation}
\subsubsection{Synthetic Data}
\begin{figure}[h!tp]
    \centering
    \makebox[\textwidth][c]{\includegraphics[width=0.9\textwidth]{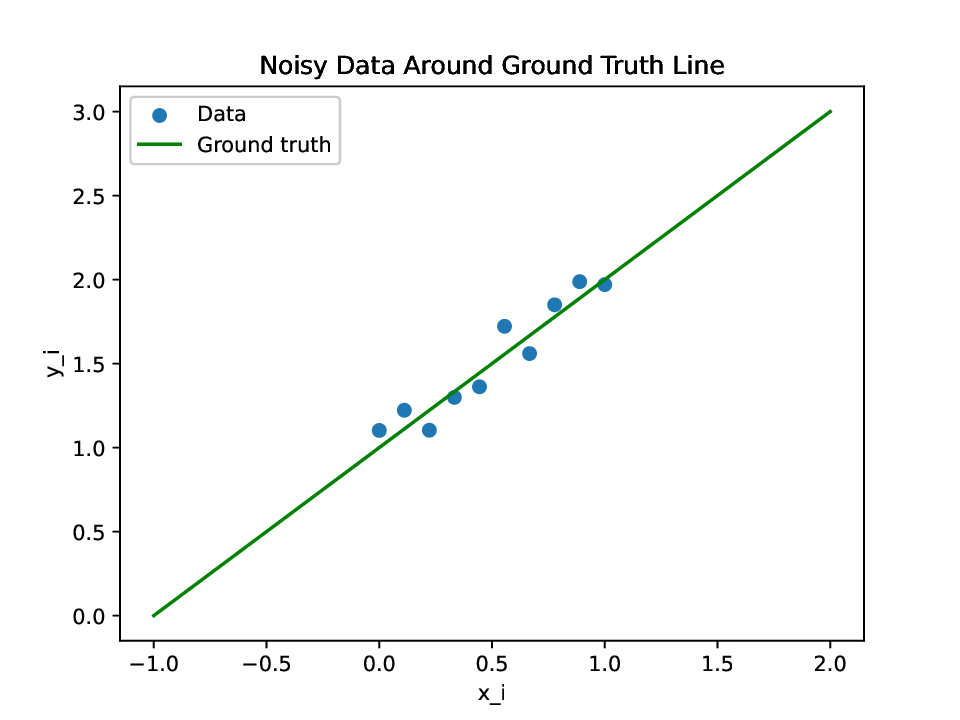}}
    \caption{Synthetic data points generated with Gaussian noise around $m(x) = x + 1$}
    \label{fig:linear_noisydata}
\end{figure}

The synthetic data is generated around the line $m(x) = ax + b$ with random Gaussian noise. We specified a mean of 0 and variance $\sigma^2$ of 0.01 to create this noise. For the ground truth equation, we set $a = 1$ and $b = 1$, making the line $m(x) = x + 1$. The synthetic data is visualized in Figure \ref{fig:linear_noisydata}.

\subsubsection{Optimal Network Architecture}\label{linear_optimalln}
We used the method outlined in Section \ref{subsec:nn} to determine a satisfactory number of layers and neurons for the PINN. The first part of this method is shown in Table \ref{tab:linear_4l_gt}. We used configurations with a constant 4 layers with varying neuron combinations. We found either 20 or 40 neurons to be most optimal because they had low MSEs calculated with ground truth function over $[-1, 2]$.

To decide between the two, we examined runtime and our knowledge of neural networks. The model with 20 neurons has significantly less runtime, and for a simple linear relationship, we know that too many neurons can cause overfitting. These factors led us to choose 20 neurons as the optimal number for 4 layers.

We next manipulated the number of layers while keeping 20 neurons fixed. Table \ref{tab:linear_20n_gt} highlights that within our configurations for 20 neurons, 3 layers resulted in the lowest average ground truth MSE over $[-1, 2]$. Both 2 and 3 layers yielded similar MSEs and had similar runtimes, but the MSE for 3 layers was smallest, so we opted for 3 layers and 20 neurons. 

However, we might not always know the ground truth function. If we instead used the average total loss as shown in Table \ref{tab:linear_4l_total}, the optimal number of neurons would be 30. While this is different than the results from Table \ref{tab:linear_4l_gt}, we see that the differences in average total loss are not that large. As a result, we can still feel confident about our choice of 20 neurons.

We also looked at the average total loss for 20 neurons and varying layers  to see how those results compared. We found that there was minimal variation between the average total loss values, as shown in Table \ref{tab:linear_20n_total}.Between the different numbers of layers for 20 neurons,  As mentioned, ground truth MSE over $[-1, 2]$ is a more reliable measure than total loss, so we continued with our choice of 3 layers.

Since our calculations for the ground truth MSE were made over the interval $[-1, 2]$ that contains data both inside and outside the training range, we also wanted to compare these results exclusively within the training range ~$[0, 1]$. In one run, as described by \ref{tab:linear_mse_int}, we compared the results of calculating error over different intervals. The optimal number of neurons did vary, illustrating the impact of different criteria in our methods. However, the MSE values are similar and performance may not vary significantly. 

Despite the discrepancy, we relied on our results from the larger interval $[-1, 2]$ because we wanted our model to generalize well to new data. The interval $[-1, 2]$ contains data both inside and outside of the training range, making it a more comprehensive range with which to calculate error.

Our final optimal network architecture for the Linear PINN was 3 layers and 20 neurons.

To access the tables and read a more detailed analysis, refer to Appendix \ref{subsec:optimallinear}.

\subsubsection{Training Process}\label{subsubsec:pinnperf}
    
\begin{figure}[h!tp]
    \centering
    \makebox[\textwidth][c]{\includegraphics[width=0.9\textwidth]{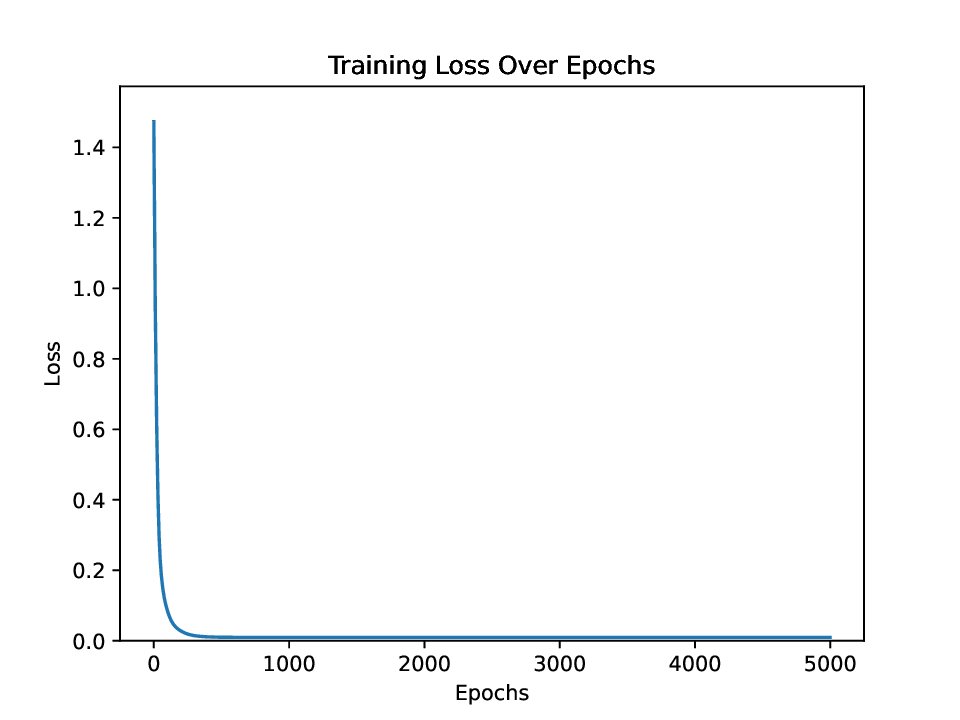}}
    \caption{Training loss for linear PINN over 5000 epochs.}
    \label{fig:linear_loss_epochs}
\end{figure}

With 3 layers, 20 neurons, $\lambda = 1.0$, and a model set-up from Section \ref{subsec:nn}, we trained the linear PINN over 5000 epochs. Figure \ref{fig:linear_loss_epochs} displays the exponential decrease of the training loss, communicating an effective training process. The average total loss over 10 runs of the model was $8.872*10^{-3}$.

\subsubsection{PINN Performance}

\begin{figure}[h!tp]
    \centering
    \makebox[\textwidth][c]{\includegraphics[width=0.9\textwidth]{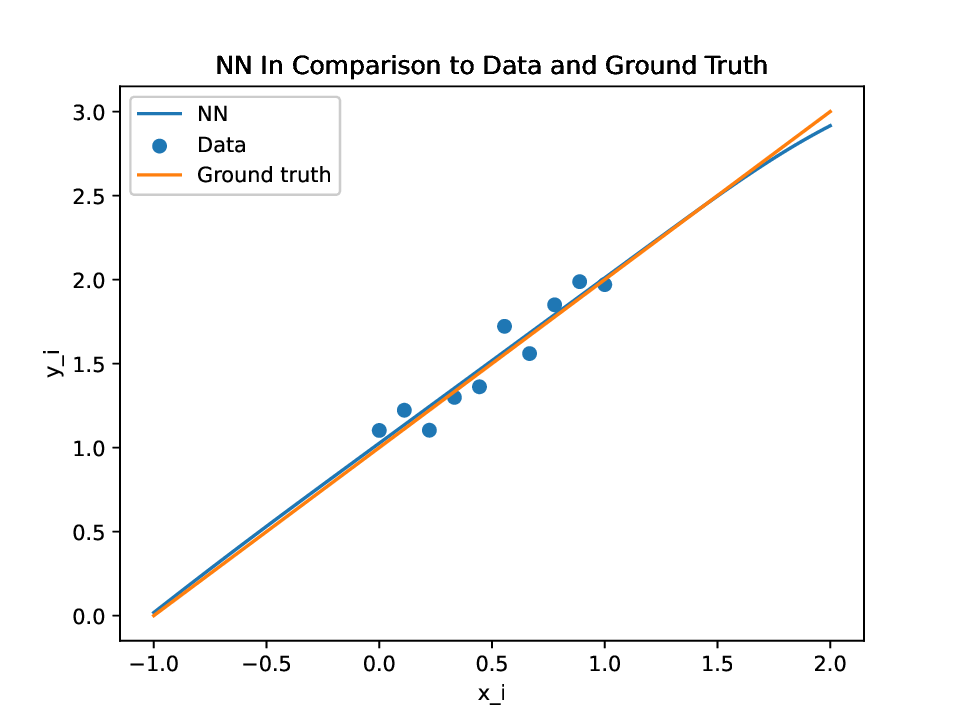}}
    \caption{Linear PINN plotted next to training data and ground truth line.}
    \label{fig:linear_nn_gt}
\end{figure}

One run of the PINN, as plotted in Figure \ref{fig:linear_nn_gt}, fit the sparse data and modeled the linear trend both inside and outside of the training range $[0, 1]$. The reason this model stayed mostly linear throughout the graph is because we minimized the residual $\frac{d^2m}{dx^2}$, feeding the neural network prior knowledge about the linear nature of the data. This neural network curved off towards $x =2.0$, which likely occurred because the model can only minimize the residual, not set it exactly to 0.

To understand the model's performance over many runs, we  calculated the average ground truth MSE of the neural network inside and outside the training range. Table \ref{tab:linear_int_perf} displays these values and communicates that the neural network performed better inside the training range than outside of it. This result makes sense because the model had never seen the data outside of that range.

\begin{table}[h!tbp]
    \centering
    \begin{tabular}{c|c}
        Interval & Avg. Ground Truth MSE\\
        \hline
        $[0, 1]$ (training range) & $1.594*10^{-3}$ \\
        \hline
        $[-1, 0]\cup[1, 2]$ (outside training range) & $1.357*10^{-2}$ \\
        \hline
        $[-1, 2]$ (overall) & $9.618*10^{-3}$ \\
    \end{tabular}
    \caption{Average ground truth MSE over different intervals for our linear PINN. The average was calculated over 10 iterations of the model.}
    \label{tab:linear_int_perf}
\end{table}

\subsubsection{Modifying Residual Strength $\lambda$}
The benefit of incorporating prior information as a residual in a neural network was outlined in Figure \ref{fig:linear_nn_gt}, but we found that the strength of the residual $\lambda$ can influence how the model fits the data. $\lambda$ can also be referred to as the regularization term. 

\begin{figure}[htbp]
    \centering
    \makebox[\textwidth][c]{\includegraphics[width=0.9\textwidth]{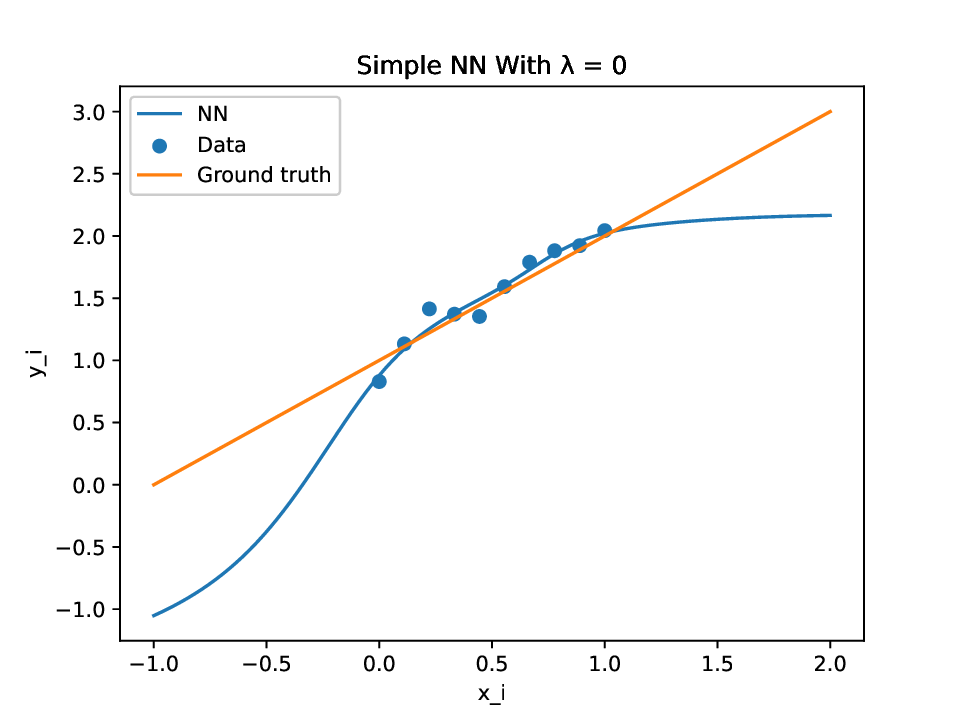}}
    \caption{Linear PINN with $\lambda = 0$, meaning the model does not consider prior knowledge about the line during training.}
    \label{fig:linear_noreg}
\end{figure}

When $\lambda = 0$, the neural network shown in Figure \ref{fig:linear_noreg} is unaware of the linear trend although it fits the data points. In fact, it values the data points so much that it overfits. This outcome was not ideal.
 
\begin{figure}
    \centering
    \begin{subfigure}{0.9\linewidth}
        \centering
        \makebox[\textwidth][c]{\includegraphics[width=\textwidth]{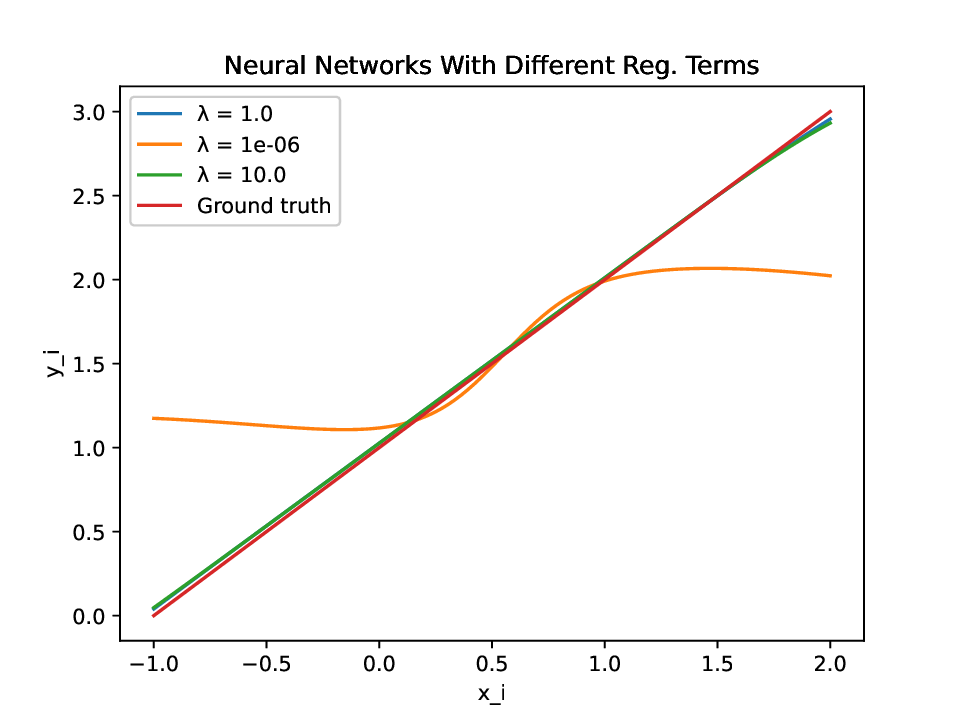}}
        \caption{Effects of manipulating the strength of the residual $\lambda$.}
        \label{fig:linear_reg_changes}
    \end{subfigure}
    \begin{subfigure}{0.9\linewidth}
        \centering
        \makebox[\textwidth][c]{\includegraphics[width=\textwidth]{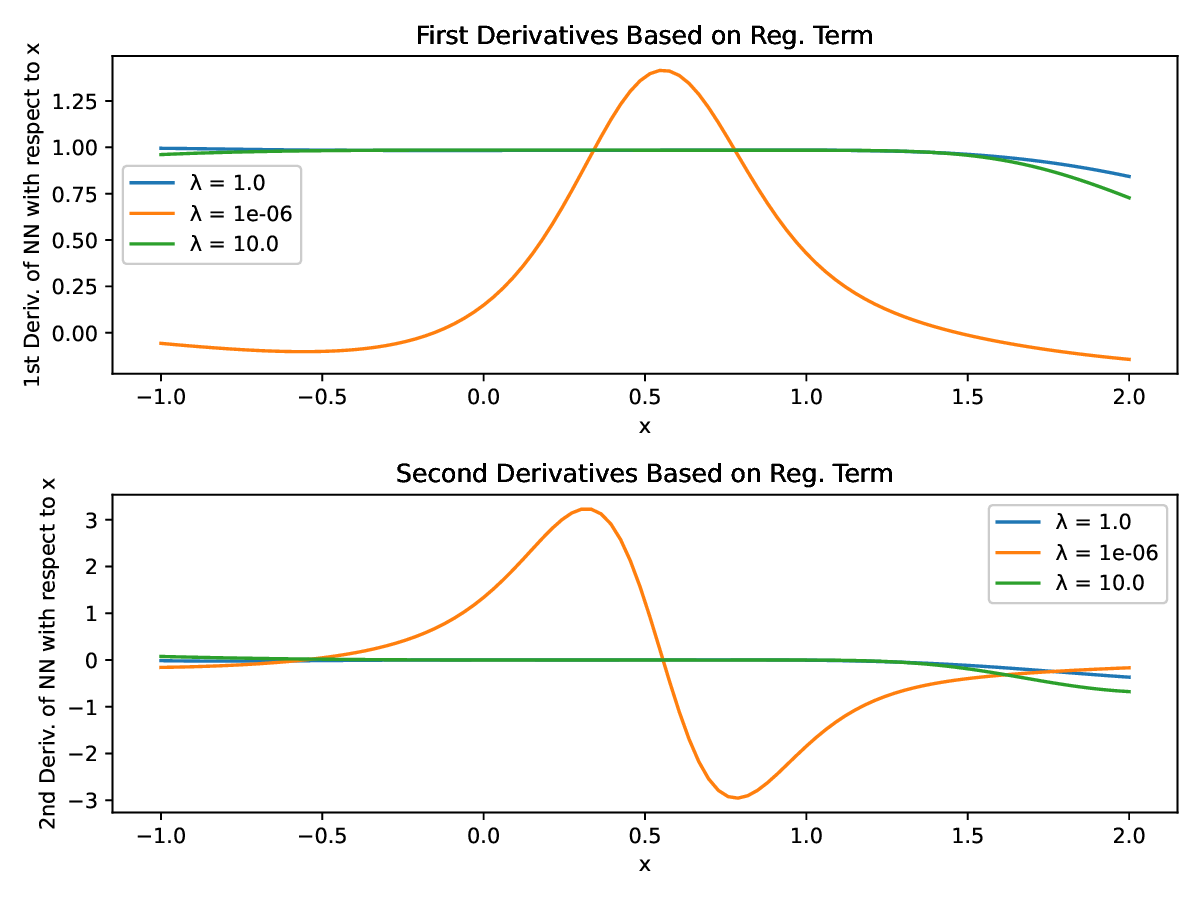}}
        \caption{First and second derivatives for models with different $\lambda$ values.}
        \label{fig:linear_reg_derivs}
    \end{subfigure}
    \caption{Neural networks and their derivatives for varying $\lambda$.}
    \label{fig:linear_reg}
\end{figure}

Then, as plotted in Figure \ref{fig:linear_reg}, we increased the value of $\lambda$ to $10^{-6}$, $1.0$, and $10.0$ such that the model increasingly incorporates the residual. 

The $\lambda = 10^{-6}$ model in Figure \ref{fig:linear_reg_changes} strayed excessively from the ground truth line when evaluated outside of $[0, 1]$; its performance indicates the small influence of the residual. Yet, the $\lambda = 1.0$ and $\lambda = 10.0$ models closely modeled the ground truth line.

When we analyzed the first and second derivatives in Figure \ref{fig:linear_reg_derivs}, we gained a deeper understanding of $\lambda$'s impact. We defined our residual as the second derivative because according to calculus, a line has a first derivative that is constant and a second derivative that is 0. Using this principle, Figure \ref{fig:linear_reg_derivs} showed us that in this particular run of the code, $\lambda = 1.0$ performed most optimally because its derivatives were the flattest. We learned from this graph that more regularization does not always lead to stronger performance; each time we train the model, the results change.

We chose $\lambda = 1.0$ as our residual strength throughout other steps of our exploration because it results in a good fit to the data. When $\lambda = 1.0$, the data loss and residual loss are valued equally, resulting in an optimal balance between fitting the data and following the larger linear trend. 

\subsubsection{Modifying the Differential Equation}
\begin{figure}[htbp]
    \centering
    \makebox[\textwidth][c]{\includegraphics[width=0.9\textwidth]{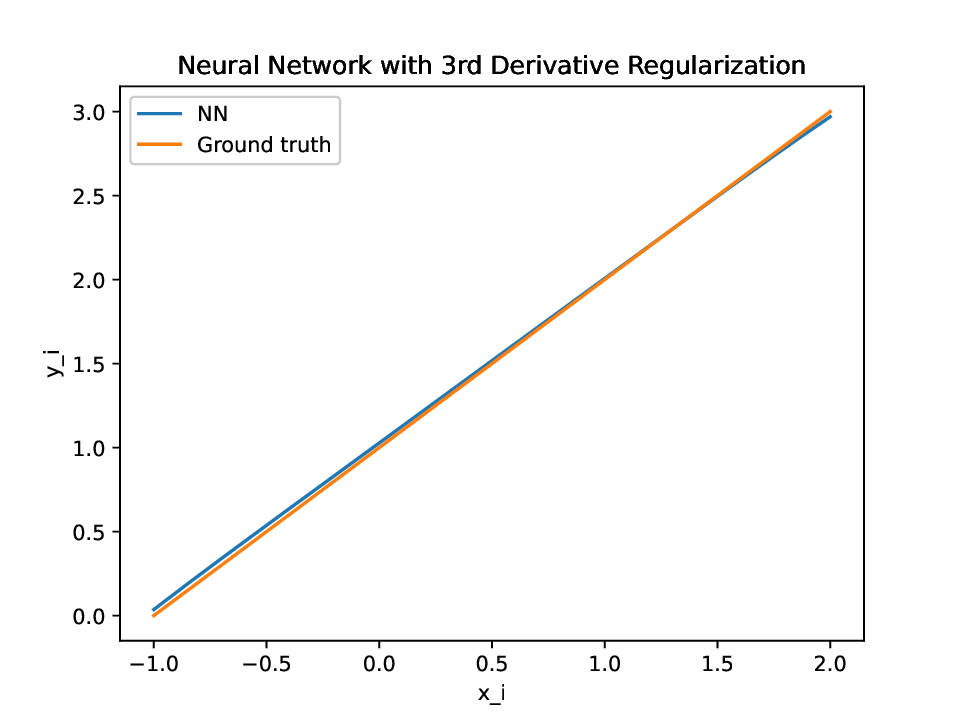}}
    \caption{Model's performance when trained with 3rd derivative residual loss.}
    \label{fig:3rdderivnet}
\end{figure}

Another area of exploration was the differential equation and consequent residual we include in the loss function. We attempted to train the model with the following differential equation that is also consistent with a line.
\begin{equation}\label{eq:linear_3rd_deriv}
    \frac{d^3m}{dx^3} = 0
\end{equation}
With the third derivative of the neural network in the loss function, the model's performance varied. In Figure \ref{fig:3rdderivnet}, the neural network fit the ground truth line closely. However, the $[-1, 2]$ ground truth MSE averaged over 10 iterations was $5.027*10^{-2}$, which is higher than the MSE of $9.618*10^{-3}$ for a model with a second derivative residual. This discrepancy in performance can be explained by the fact that a parabola shares the relationship described by Equation \ref{eq:linear_3rd_deriv}, so the model may not always retain the linear trend.

\subsection{Quadratic PINN}
We set up our first PINN to learn from noisy synthetic data that is consistent with the differential equation belonging to a quadratic function: \begin{equation}
\frac{d^3m}{dx^3} = 0
\end{equation}

\subsubsection{Synthetic Data}
\begin{figure}[h!tp]
    \centering
    \makebox[\textwidth][c]{\includegraphics[width=0.9\textwidth]{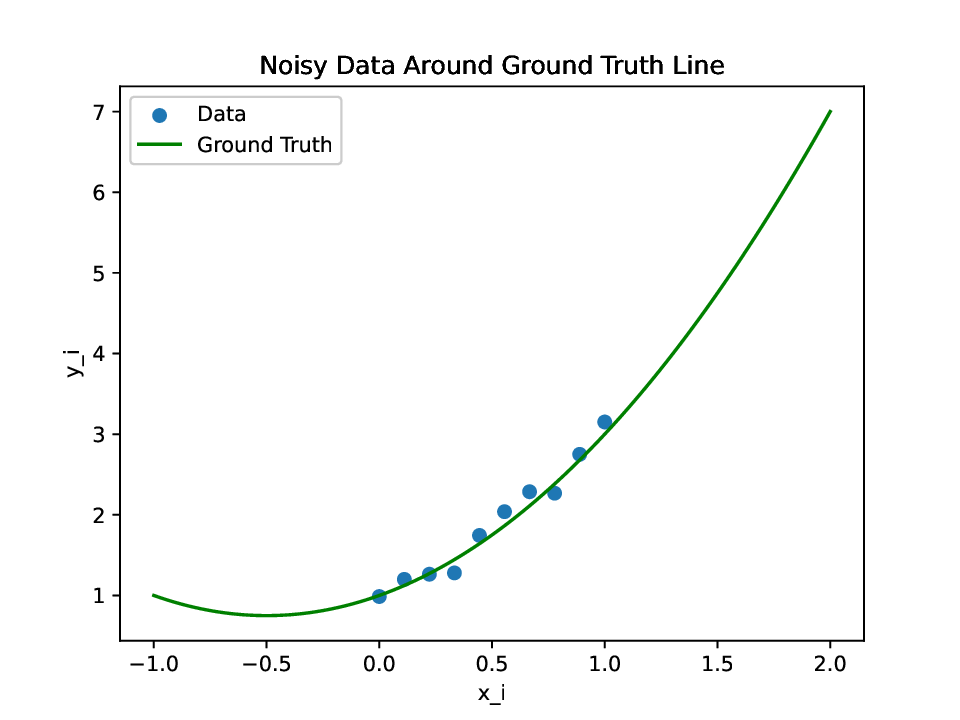}}
    \caption{Synthetic data points generated with Gaussian noise around $m(x) = x^2 + x + 1$}
    \label{fig:quad_noisydata}
\end{figure}

The synthetic data is generated around the line $m(x) = ax^2 + b + c$ with random Gaussian noise. We specified a mean of 0 and variance $\sigma^2$ of 0.01 to create this noise. For the ground truth equation, we set $a = 1$, $b = 1$, and $c = 1$, making the line $m(x) = x^2 + x + 1$. The synthetic data is visualized in Figure \ref{fig:quad_noisydata}.

\subsubsection{Optimal Network Architecture}\label{quad_optimalln}
We used the same method outlined in Section \ref{subsec:nn} as before to determine a satisfactory number of layers and neurons for the PINN. The first part of this method is shown in Table \ref{tab:quad_4layers}. It entails 5 configurations with 4 layers and varying neurons.

When holding layers constant at 4, we observed that 30 neurons yields the best result. Unlike our linear model, the difference between the ground truth MSE for 30 neurons and that of any other configuration was fairly vast, so we felt more confident in our choice.

Then, as shown in Table \ref{tab:quad_30neurons} we kept the number of neurons constant at 30 and changed the number of layers in our model. We found that 5 layers would be the most optimal for 30 neurons because it had the lowest average ground truth MSE. Thus, our optimal layer count was 5 layers.

We also considered the impact of a different criterion for our method, such as using the average total loss instead of the average ground truth MSE. The average total loss is calculated during training using the data and the residual instead of the ground truth function. The average loss for the different neuron configurations is shown in Table \ref{tab:quad_4layers_total}. We see there that 30 neurons yield the lowest average total loss for a fixed number of layers of 4. This result using the total loss was consistent with the results using the ground truth function. We can see this similarity when comparing Tables \ref{tab:quad_4layers} and \ref{tab:quad_4layers_total}.

We also examine total loss while keeping the neuron per layer count constant at 30 and varying the layer count. In Table \ref{tab:quad_30neurons_total}, we see that 5 layers is the optimal configuration with a low average total loss. 

Another consideration in our method is the interval we use to calculate ground truth MSE. In Table \ref{tab:quad_4layers_interval}, we examined the difference between calculating ground truth MSE inside the interval $[0, 1]$ and $[-1, 2]$ during one run to determine how the results of those methods compare. In that run, the optimal number of neurons and layers stayed the same for both intervals at 30 neurons per layer and 5 layers. Unlike our linear model, the choice of interval here did not impact the final network architecture.

We used the ground truth MSE over $[-1, 2]$ to compare our model with the actual trends instead of the noisy data. As explained in Section \ref{subsec:optimallinear}, we wanted to use the interval $[-1, 2]$ in our tests to capture performance outside of the training interval.

Given these considerations, the final optimal network architecture for the Linear PINN was 5 layers and 30 neurons.

To access the tables and read a more detailed analysis, refer to Appendix \ref{subsec:optimalquad}.

\subsubsection{Training Process}\label{subsubsec:quad_train}
    
\begin{figure}[h!tp]
    \centering
    \makebox[\textwidth][c]{\includegraphics[width=0.9\textwidth]{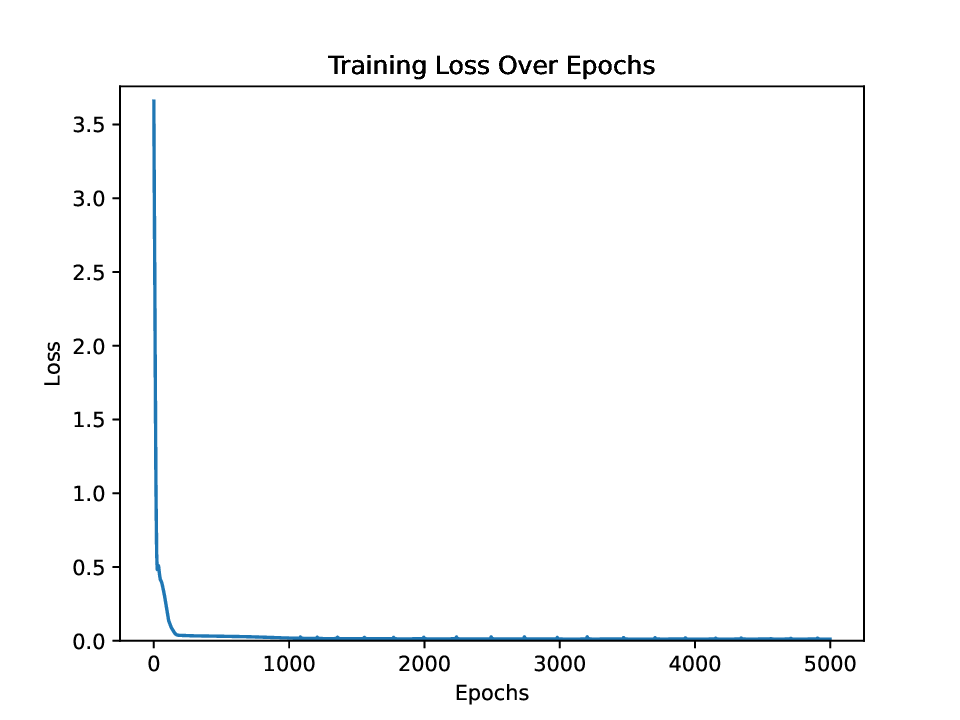}}
    \caption{Training loss for quadratic PINN over 5000 epochs.}
    \label{fig:quad_loss_epochs}
\end{figure}

With 5 layers, 30 neurons, $\lambda = 1.0$, and a model set-up from Section \ref{subsec:nn}, we trained the quadratic PINN over 5000 epochs. Figure \ref{fig:quad_loss_epochs} displays the exponential decrease of the training loss, communicating an effective training process. The average total loss over 10 runs of the model was $5.671*10^{-3}$.

\subsubsection{PINN Performance}

\begin{figure}[h!tp]
    \centering
    \makebox[\textwidth][c]{\includegraphics[width=0.9\textwidth]{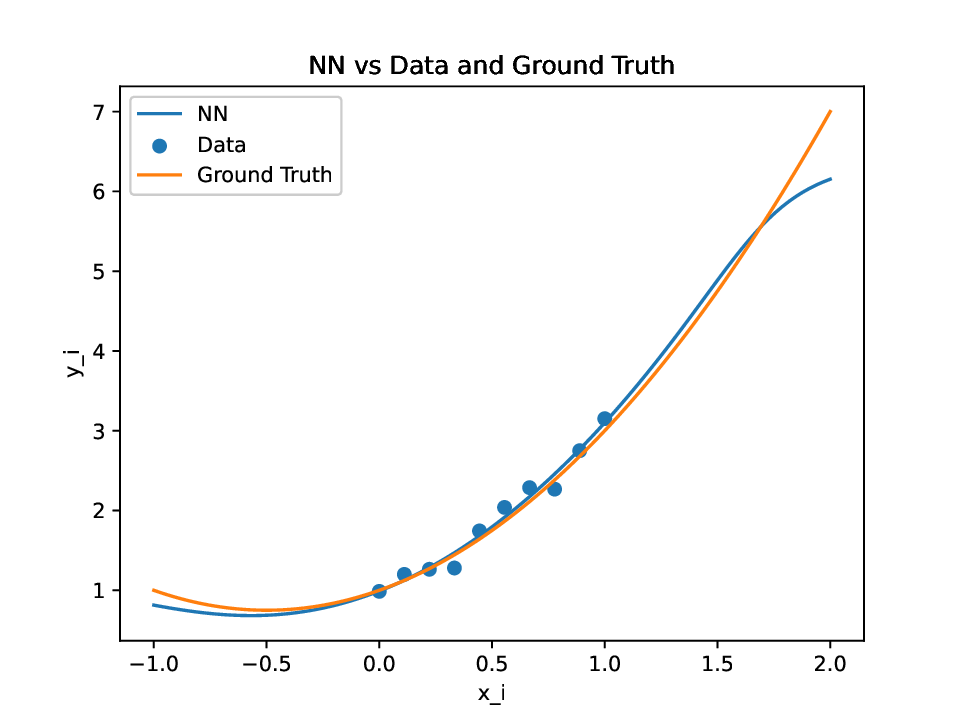}}
    \caption{Quadratic PINN plotted next to training data and ground truth line.}
    \label{fig:quad_nn_gt}
\end{figure}

One iteration of the PINN, as plotted in Figure \ref{fig:quad_nn_gt}, fits the sparse data and models the quadratic trend both inside and outside the training interval $[0, 1]$. Our PINN remained on the parabola throughout the graph because we minimized the residual $\frac{d^3m}{dx^3}$ which fed the neural network prior knowledge about the quadratic nature of the data. This neural network curved off towards $x \approx 1.7$. This occurs because while the model attempts to minimize the residual to 0, it can't get the residual to equal exactly zero. The model's inability to perfectly minimize the residual allows error to be added to our system.

To understand the model's performance over many runs, we  calculated the average ground truth MSE of the neural network inside and outside the training range just like the Linear PINN. Table \ref{tab:quad_int_perf} displays these values and communicates that the neural network performed better inside the training interval than outside of it. This result makes sense because the model had never seen the data outside that range. However, due to the strength of the residual that we influenced earlier, the model still follows trends in data outside the training interval as well.

\begin{table}[h!tbp]
    \centering
    \begin{tabular}{c|c}
        Interval & Avg. Ground Truth MSE\\
        \hline
        $[0, 1]$ (training range) & $1.495*10^{-1}$ \\
        \hline
        $[-1, 0]\cup[1, 2]$ (outside training range) & $2.702*10^{-3}$ \\
        \hline
        $[-1, 2]$ (overall) & $2.215*10^{-1}$ \\
    \end{tabular}
    \caption{Average ground truth MSE over different intervals for our quadratic PINN. The average was calculated over 10 iterations of the model.}
    \label{tab:quad_int_perf}
\end{table}
    
\subsubsection{Modifying Residual Strength $\lambda$}
Just like our Linear PINN, we recognize the changing the $\lambda$ parameter value in our model can significantly change how well our model performs. 

\begin{figure}[htbp]
    \centering
    \makebox[\textwidth][c]{\includegraphics[width=0.9\textwidth]{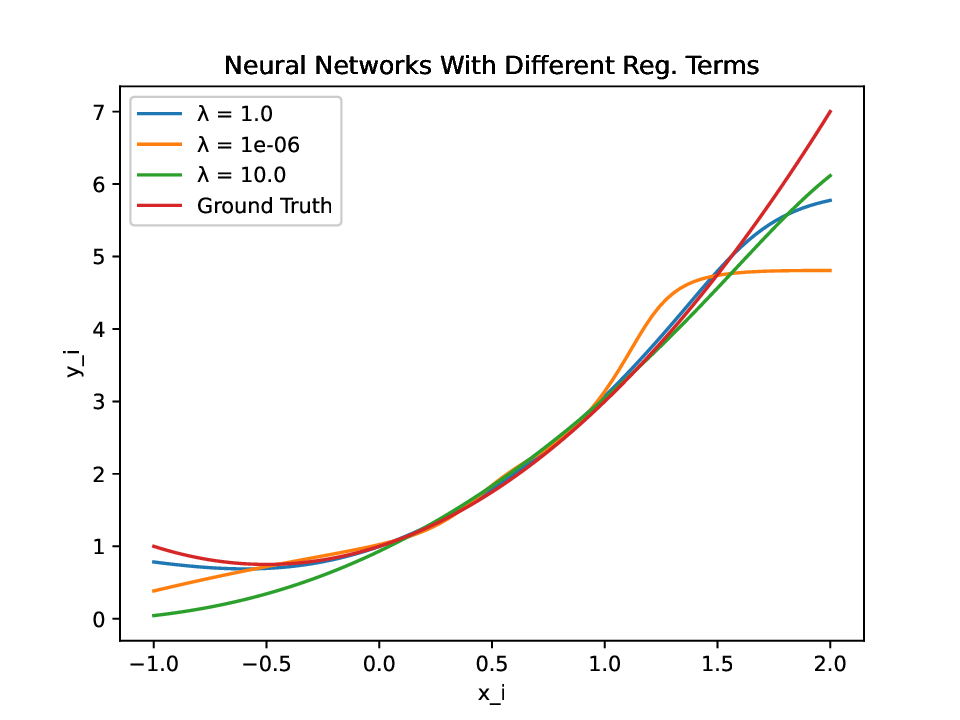}}
    \caption{Quadratic PINN with $\lambda = 1.0, 10.0, 10^{-6}$, which means prior information effects the model different based on the parameter value given.}
    \label{fig:quad_reg}
\end{figure}

When $\lambda = 0$ we have already seen that the model isn't satisfactory outside the training interval, and in Figure \ref{fig:quad_reg} we train the model with different $\lambda$ values to illustrate how influencing the strength of our residual can influence the effectiveness of our model.

As Figure \ref{fig:quad_reg} shows, the model with $\lambda = 10^{-6}$ strays from the ground truth when it is evaluated outside the training interval. Evidently, a lower lambda value decreases the influence of the residual and makes our model ineffective, which is consistent with our findings from the Linear PINN.

Additionally, Figure \ref{fig:quad_reg} shows that a model with $\lambda = 10.0$ and just like we observed with the Linear PINN, our residual strength can be too powerful which causes the model to become more ineffective. Finally we came to the conclusion that the model with $\lambda = 1.0$ yields the best result when compared to the ground truth line.

\begin{figure}[htbp]
    \centering
    \makebox[\textwidth][c]{\includegraphics[width=0.9\textwidth]{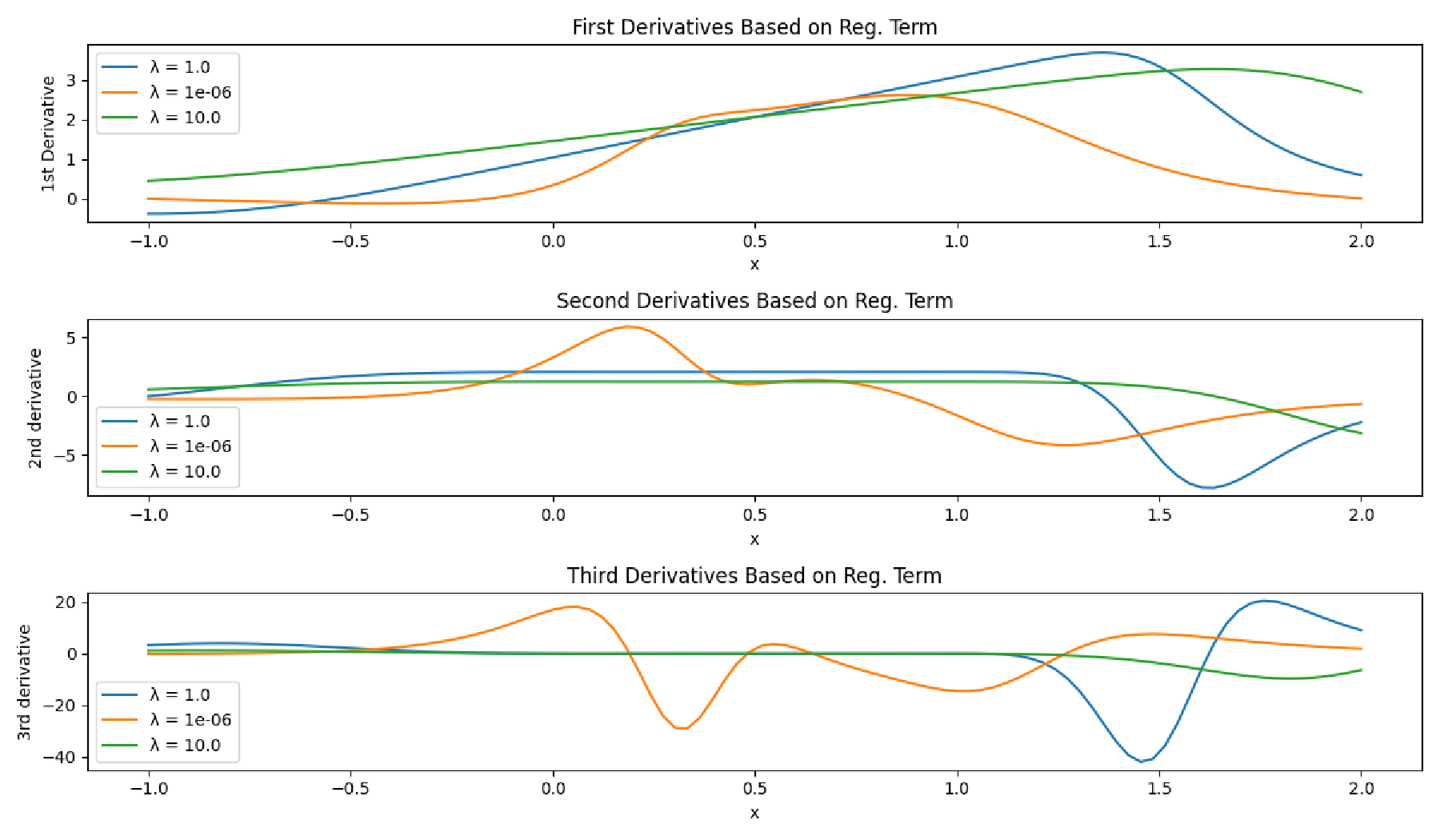}}
    \caption{Quadratic PINN with $\lambda = 1.0, 10.0, 10^{-6}$, and the derivatives with each $\lambda$ value.}
    \label{fig:quad_deriv}
\end{figure}

Figure \ref{fig:quad_deriv} shows the different derivatives at different $\lambda$ values, which provides insight on how well the model is able to minimize the residual based on the varying $\lambda$ value.

Just like \ref{fig:quad_reg} evidently the model with $\lambda = 10^{-6}$ performs the worst on each derivative as it isn't highly influenced by the residual.

\subsubsection{Modifying the Differential Equation}
Another area of exploration was the differential equation and consequent residual we include in the loss function. We attempted to train the model with the following differential equation that is also consistent with a line.
\begin{equation}\label{eq:quad_4th_deriv}
    \frac{d^4m}{dx^4} = 0
\end{equation}
With the fourth derivative of the neural network in the loss function, the model's performance varied. Just like the Linear PINN we can continue to change which derivative we minimize and that will change how our model performs.

\begin{figure}[htbp]
    \centering
    \makebox[\textwidth][c]{\includegraphics[width=0.9\textwidth]{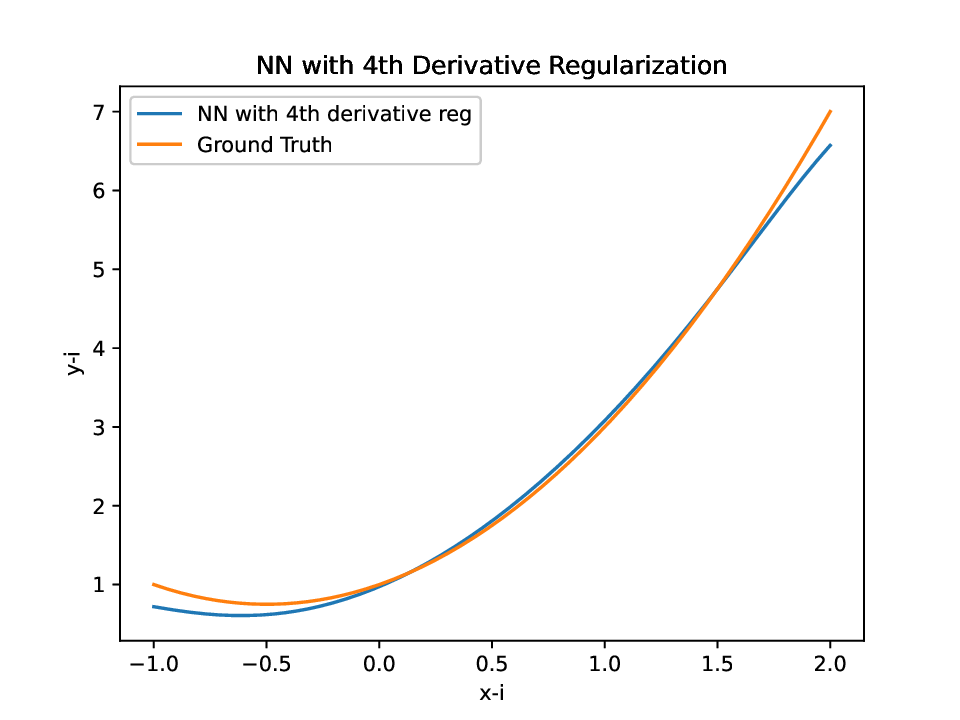}}
    \caption{Model's performance when trained with 3rd derivative residual loss.}
    \label{fig:quad_deriv4}
\end{figure}

Figure \ref{fig:quad_deriv4} illustrates training our model with \ref{eq:quad_4th_deriv} while keeping all the other model properties constant.

In Figure \ref{fig:quad_deriv4}, the neural network fits the ground truth line closely. However, the $[-1, 2]$ ground truth MSE averaged over 10 iterations was $1.648*10^{-1}$, which is higher than the MSE of $1.495*10^{-1}$ for a model with a third derivative residual. This discrepancy in performance can be explained by the fact that a parabola shares the relationship described by Equation \ref{eq:quad_4th_deriv}, so the model may not always retain the quadratic trend.

\subsection{Heat Equation PINN}

\subsubsection{Synthetic Data}
\begin{figure}[htbp]
    \centering
    \makebox[\textwidth][c]{\includegraphics[width=0.9\textwidth]{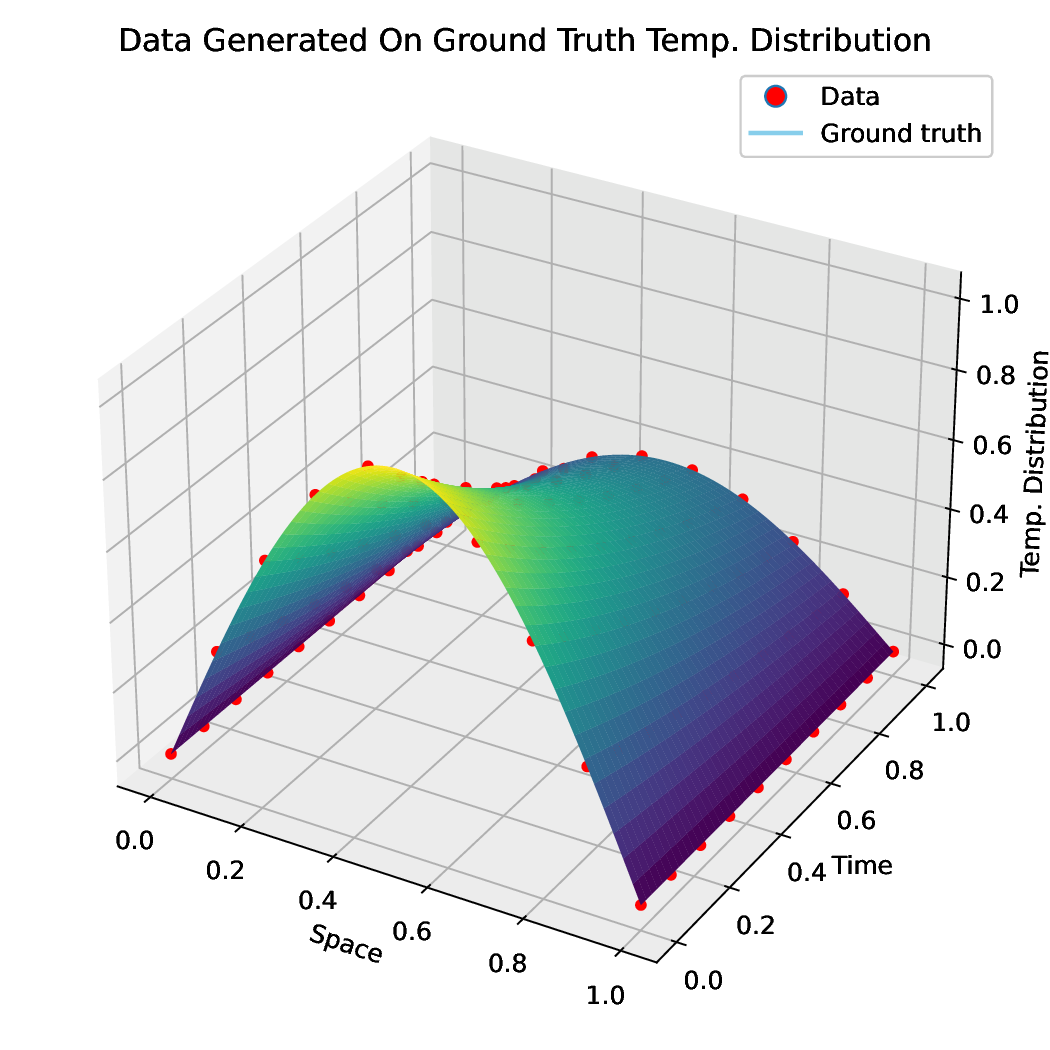}}
    \caption{Data generated on ground truth surface of temperature distribution across space and time.}
    \label{fig:heat_3d_datagt}
\end{figure}

We generated the synthetic data on the ground truth surface $u(x, t) = e^{-\pi^2Dt}\sin{\pi x}$, as shown in Figure \ref{fig:heat_3d_datagt}. To solve the forward problem, we decided on $D = 0.1$ as the thermal diffusivity parameter. To generate the data, we chose 0.1 for $\Delta x$ and $\Delta t$, which creates 100 points for the overall dataset.

\begin{figure}
    \centering
    \begin{subfigure}{0.8\linewidth}
        \centering
        \makebox[\textwidth][c]{\includegraphics[width=0.8\textwidth]{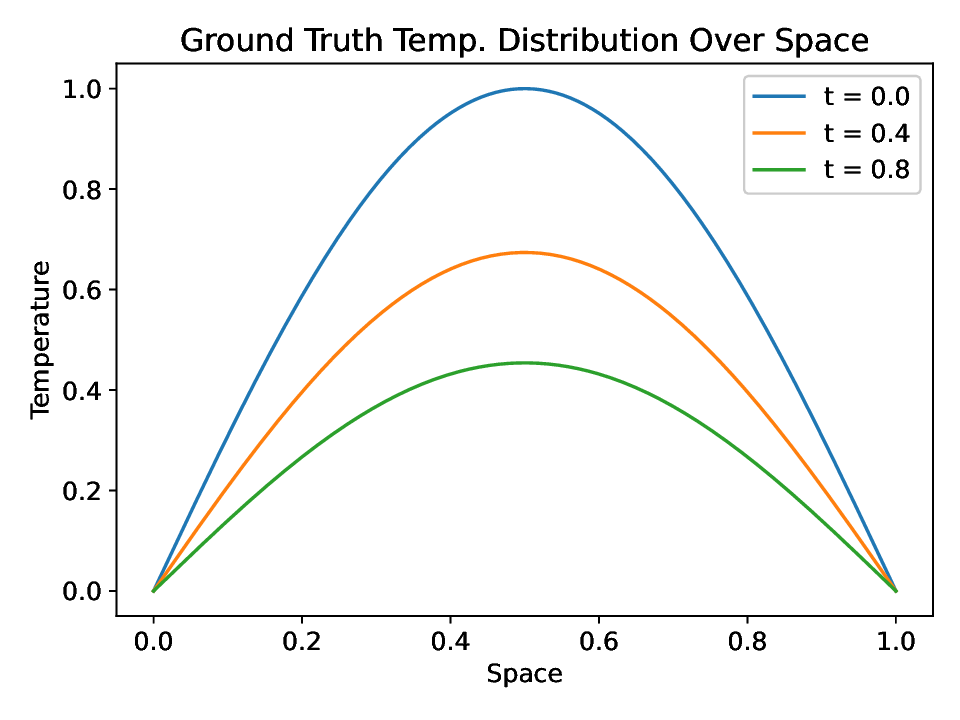}}
        \caption{Ground truth temperature distribution over space at different t values.}
        \label{fig:heat_temp_space}
    \end{subfigure}
    \begin{subfigure}{0.8\linewidth}
        \centering
        \makebox[\textwidth][c]{\includegraphics[width=0.8\textwidth]{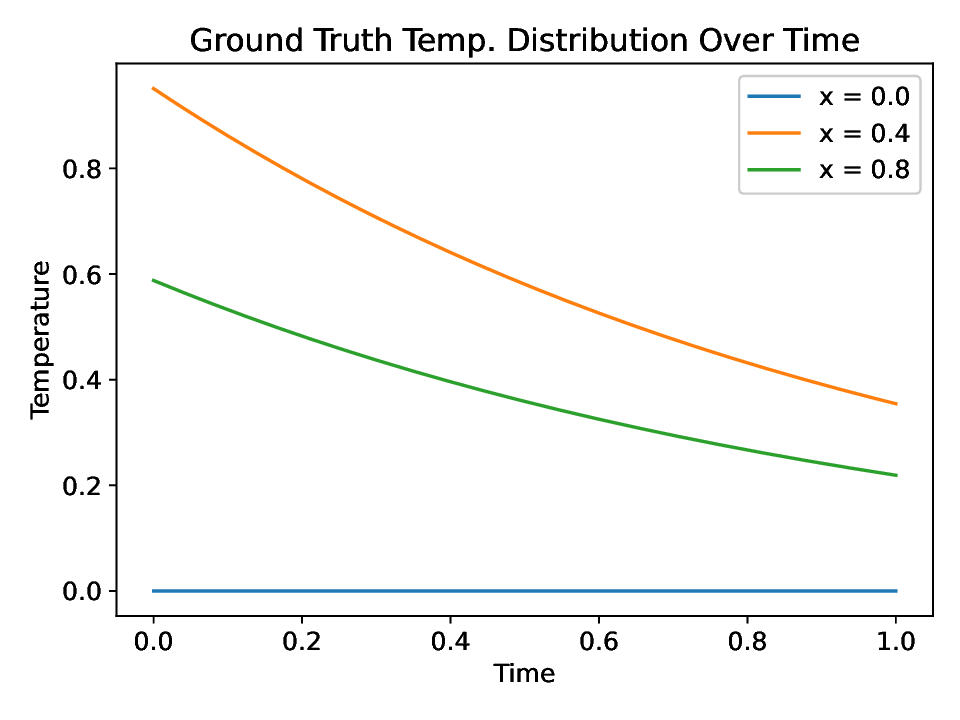}}
        \caption{Ground truth temperature distribution over time at different x values.}
        \label{fig:heat_temp_time}
    \end{subfigure}
    \caption{Ground truth or exact solution of Equation \ref{eq:heat} plotted in 2D.}
    \label{fig:heat_2d}
\end{figure}

To get a closer view of the ground truth surface, we can plot it in 2D as shown in Figure \ref{fig:heat_2d}.

\subsubsection{Optimal Network Architecture}\label{subsubsec:heat_optimalln}

We again used the method outlined in Section \ref{subsec:nn} to determine a satisfactory configuration of layers and neurons for our PINN. 

We began by setting 4 layers constant and creating 5 different models that vary the number of neurons in that layer. The configurations we first tested were 4 layers and 10, 20, 40, 60, and 80 neurons in each of those layers. Then, we trained the models, calculated the average ground truth MSE on $[0, 1]$ over 5 iterations, and chose the model with the smallest MSE value. The results are show in Table \ref{tab:heat_4l_gt}, concluding that 80 neurons is optimal given 4 fixed layers. 

The next step was to set up 5 models with a constant 80 neurons per layer and a varying number of layers. The variations were 2, 3, 4, 5, and 6 total layers, once again calculating the average ground truth MSE after training these models. The averages, as displayed in Table \ref{tab:heat_80n_gt}, indicate that 5 layers is the optimal number for 80 neurons. 

5 layers and 80 neurons per layer was our optimal network architecture for our Heat Equation PINN.

However, it's important to address a limitation of this process, which is that we rely on knowing the ground truth function. For this reason, we also calculated the average total loss of those models over 5 iterations to see how the results compared. Both Table \ref{tab:heat_4l_total} and Table \ref{tab:heat_80n_total} indicate that even if we used the model's total loss to decide on the network architecture, we would get the same result of 5 layers and 80 neurons per layer.

In Sections \ref{linear_optimalln} and \ref{quad_optimalln}, we also examined a scenario where the ground truth MSE was calculated over a larger interval $[-1, 2]$. We want to note that our Heat Equation PINN can only be evaluated on $[0, 1]$ because of the way boundary and initial conditions are enforced. Thus, the only way we calculated ground truth MSE was on the interval $[0, 1]$.

To access the tables and read a more detailed analysis, refer to Appendix \ref{subsec:optimalheat_appendix}.

\subsubsection{Training Process}
\begin{figure}[h!tp]
    \centering
    \makebox[\textwidth][c]{\includegraphics[width=0.9\textwidth]{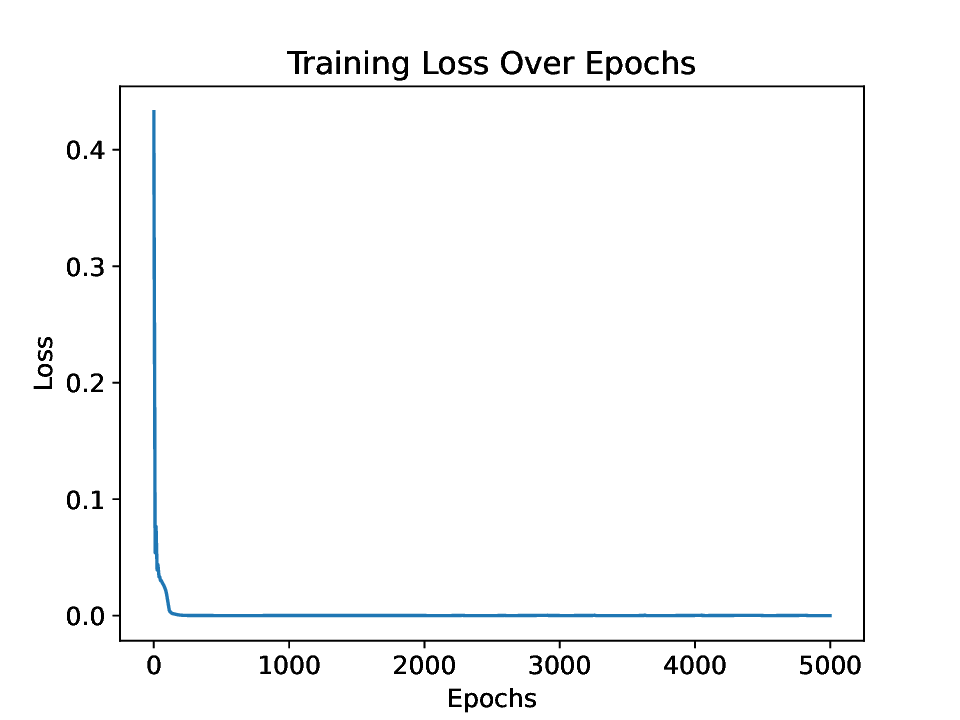}}
    \caption{Training loss for Heat Equation PINN over 5000 epochs.}
    \label{fig:heat_loss_epochs}
\end{figure}

With 5 layers, 80 neurons, $\lambda = 1.0$, and a model set-up from Section \ref{subsec:nn}, we trained the Heat Equation PINN over 5000 epochs. Figure \ref{fig:heat_loss_epochs} displays the exponential decrease of the training loss, communicating an effective training process. The average total loss over 10 runs of the model was \{Insert Loss after iterations\}.

\subsubsection{PINN Performance}

\begin{figure}[h!tp]
    \centering
    \makebox[\textwidth][c]{\includegraphics[width=0.9\textwidth]{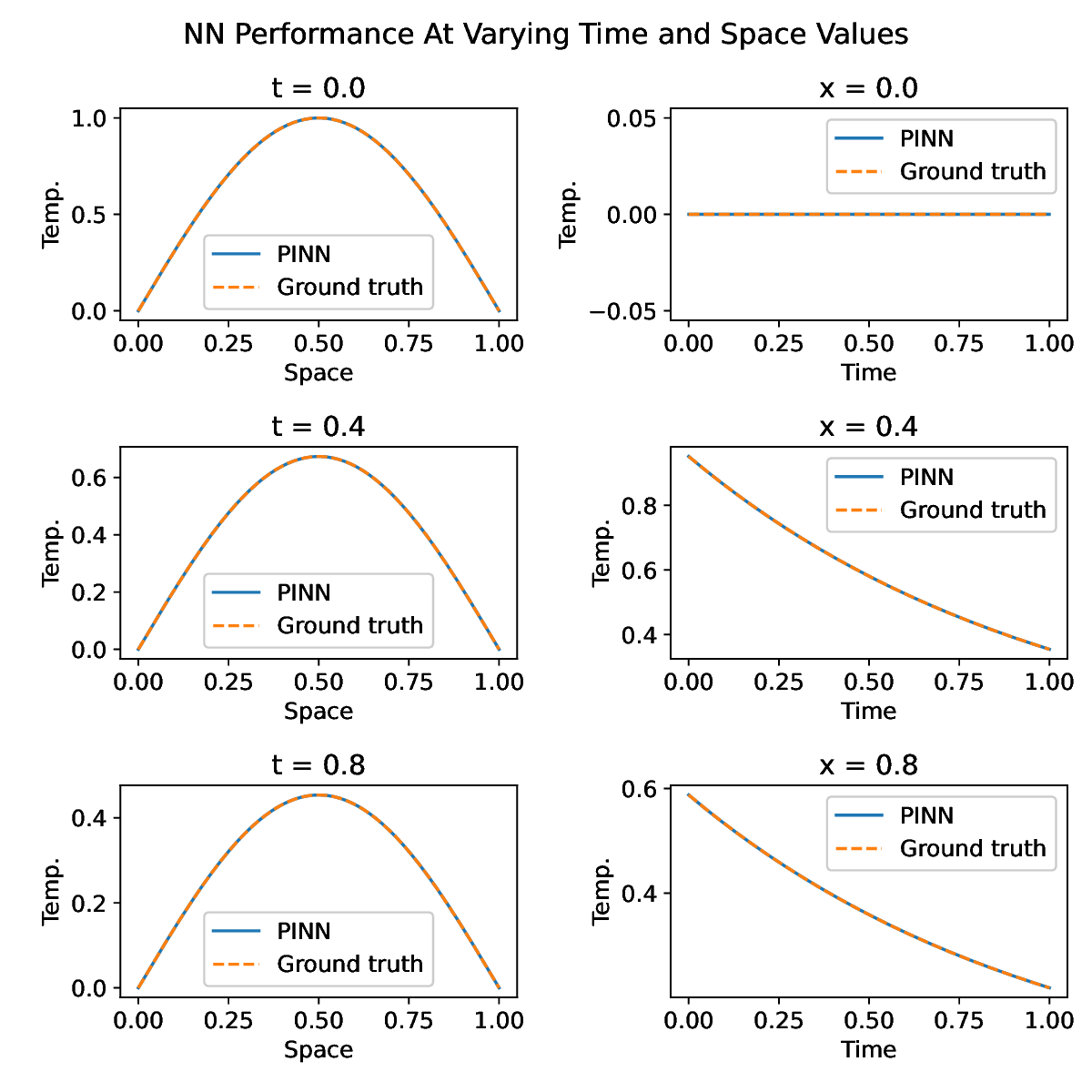}}
    \caption{Performance of neural network at varying time and space values versus the ground truth line. First with time and space at $0$. Second with time and space at $0.4$. Third with time and space at $0.8$.}
    \label{fig:heat_pinn_time_space}
\end{figure}

One iteration of the PINN as shown in Figure \ref{fig:heat_pinn_time_space} attempts to fit the ground truth line shown in blue. We plot multiple graphs at time and space values. First, we plot the network at time and space at $0$ where we enforced the boundary condition to ensure the network is at $0$ when time is $0$. 

Evidently, Figure \ref{fig:heat_pinn_time_space} shows that the PINN is able to fit the ground-truth line extremely effectively.

\begin{figure}[h!tp]
    \centering
    \makebox[\textwidth][c]{\includegraphics[width=0.9\textwidth]{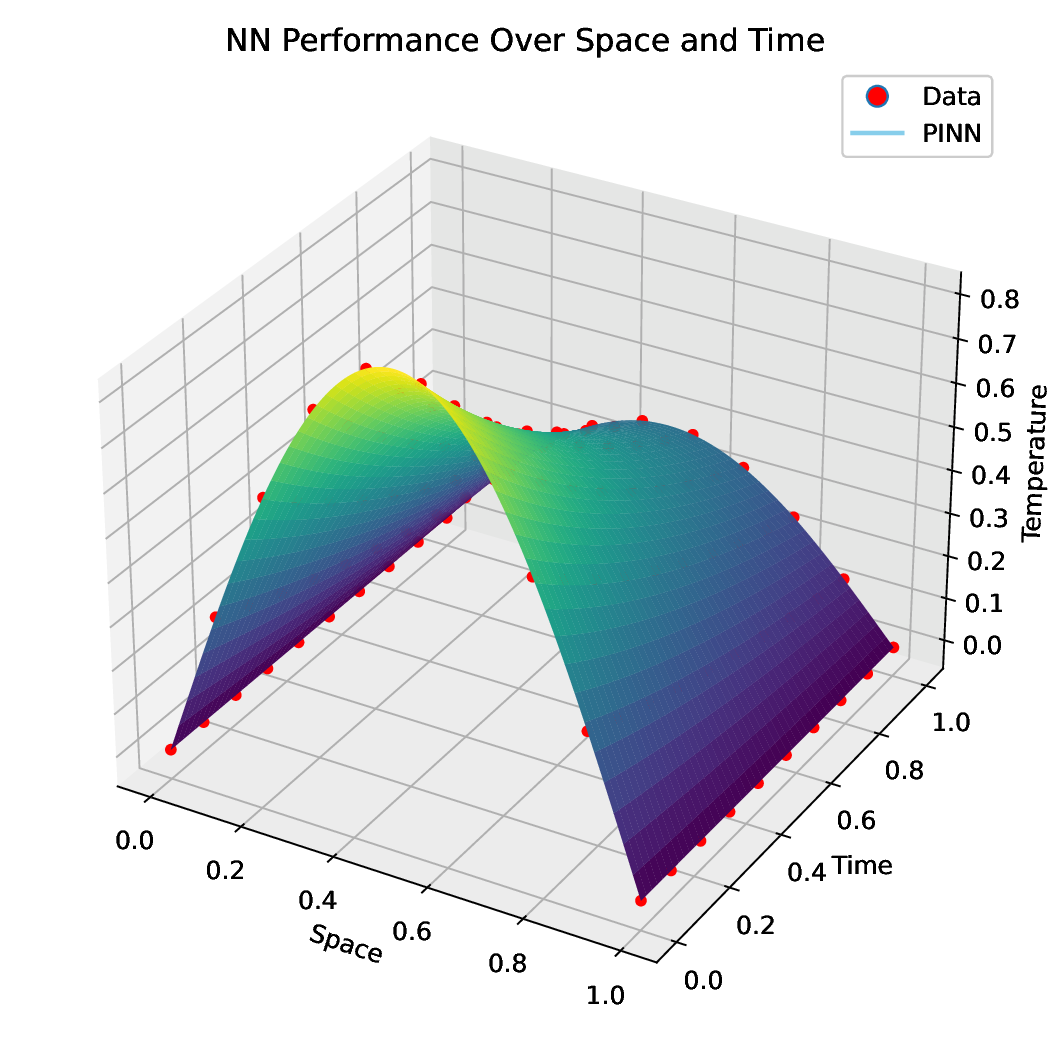}}
    \caption{Performance of neural network versus the ground truth line with 3 axes. We plot Temperature Distribution over Space and Time. The red dots denote synthetic data points and the 3D surface denotes our PINN.}
    \label{fig:heat_pinn_data}
\end{figure}

We can use another method of plotting the PINN over the ground-truth function by using a 3D graph. Figure \ref{fig:heat_pinn_data} uses the red dots as the synthetic data points that we generated, and the surface as our PINN. Using this figure, we can observe the changes in temperature distribution over all the data points in our dataset at all times in our dataset.

Just like Figure \ref{fig:heat_pinn_time_space} the PINN fits the synthetic data points extremely well.

To further understand the accuracy of our model, we can look at the average MSE and Total Loss of our model over 10 iterations. Table \ref{tab:heat_forward_iters} examines the two types of loss and the time each iteration took.

\begin{table}[htbp]
    \centering
    \begin{tabular}{c|c}
        Metric & Average\\
        \hline
        $[0, 1]$ Ground Truth MSE & $2.720*10^{-6}$ \\
        \hline
        Total Loss & $1.472*10^{-5}$ \\
        \hline
        Time (sec.) & $70.886$\\
    \end{tabular}
    \caption{Average MSE, Total Loss, and Time for all 10 iterations of the Heat Equation PINN}
    \label{tab:heat_forward_iters}
\end{table}

Although our model took more time to train as the data was more complex. Our average MSE and Total Loss are extremely low, and we can conclude that our model performed effectively.

\subsubsection{Examining Residuals}

As described in \ref{subsec:nn} the goal of our PINN is to minimize the residual. In this case, the residual that we want to minimize is

\begin{equation}\label{eq:heat_residual}
    \frac{\partial u}{\partial t} - k \frac{\partial^2 u}{\partial x^2}
\end{equation}

\begin{figure}[h!tp]
    \centering
    \makebox[\textwidth][c]{\includegraphics[width=0.9\textwidth]{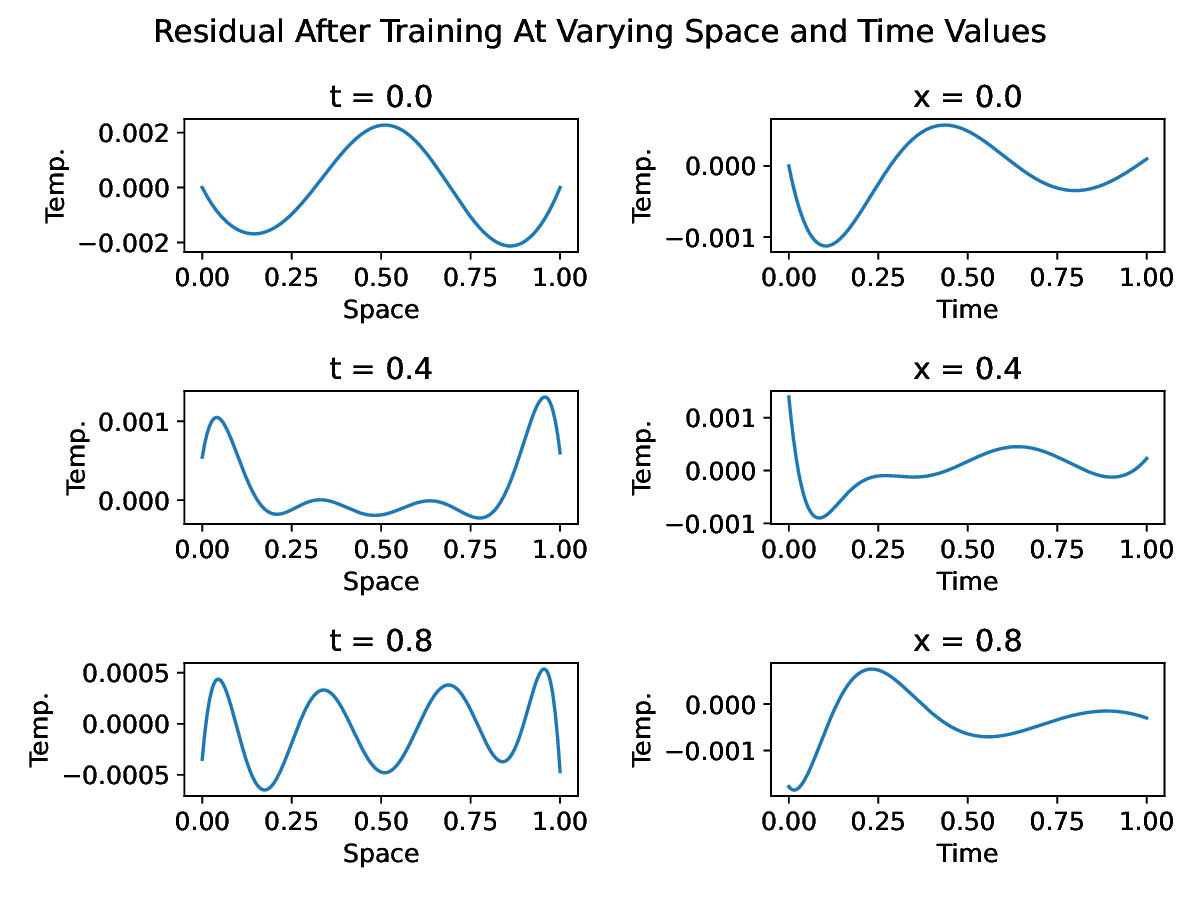}}
    \caption{Residual from the heat equation plotted at varying space and time values.}
    \label{fig:heat_residual}
\end{figure}

In Figure \ref{fig:heat_residual} we observe the residual of our PINN at varying space and time values. For all the graphs, the range of the function defining the residual is minimal. Therefore, even if the residual seems chaotic intuitively; the graph portrays the residual at near-zero values.

As the value of time gets larger, the residual begins to smooth out, which is evident as the range of the residual gets smaller. Our PINN is also effective as minimizing the residual as shown by Figure \ref{fig:heat_residual} as well.

\subsubsection{Influence of Thermal Diffusivity (D)}
The thermal diffusivity value (D) in Equation \ref{eq:heat}, which measures how quickly temperature concavity, is smoothed out can greatly affect our model. Additionally, the density of our grid which can be determined by the value of $\Delta x$ and $\Delta t$ also affects our model based on the thermal diffusivity value given. 

\begin{figure}[htbp]
\begin{subfigure}{.5\textwidth}
  \centering
  \includegraphics[width=\linewidth]{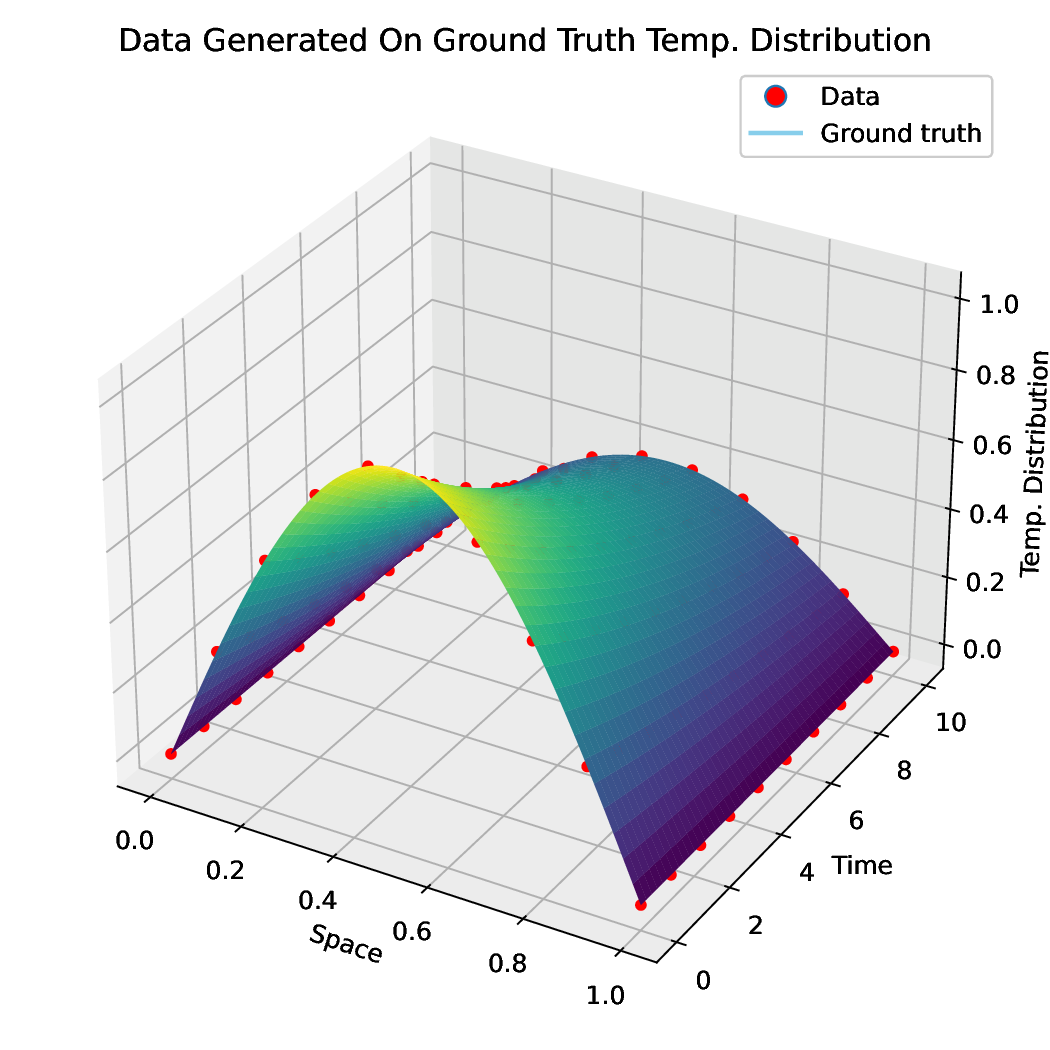}  
  \caption{Data with D = $0.01$}
  \label{fig:temp_space_0.01_3d}
\end{subfigure}
\begin{subfigure}{.5\textwidth}
  \centering
  \includegraphics[width=\linewidth]{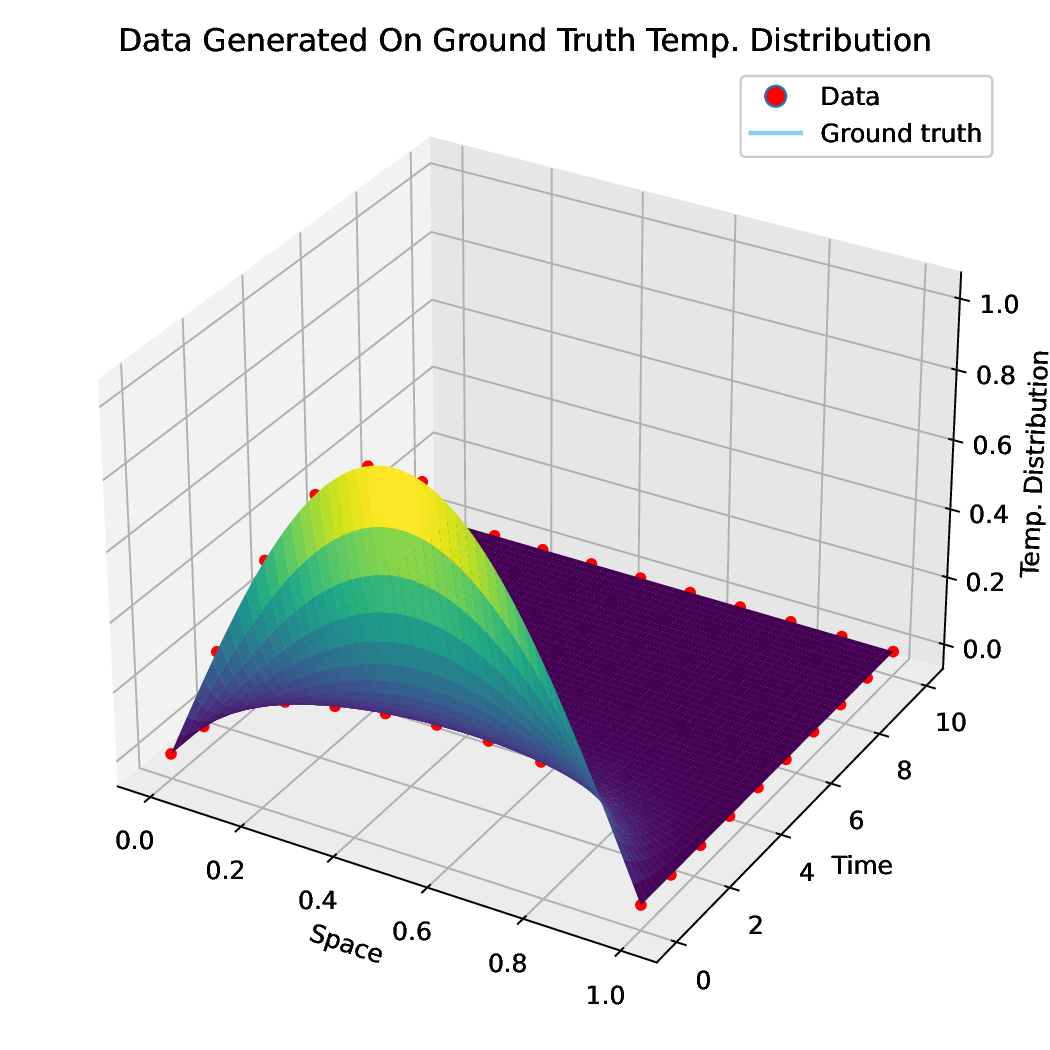}  
  \caption{Data with D = $0.1$}
  \label{fig:temp_space_0.1_3d}
\end{subfigure}
\begin{subfigure}{.5\textwidth}
  \centering
  \includegraphics[width=\linewidth]{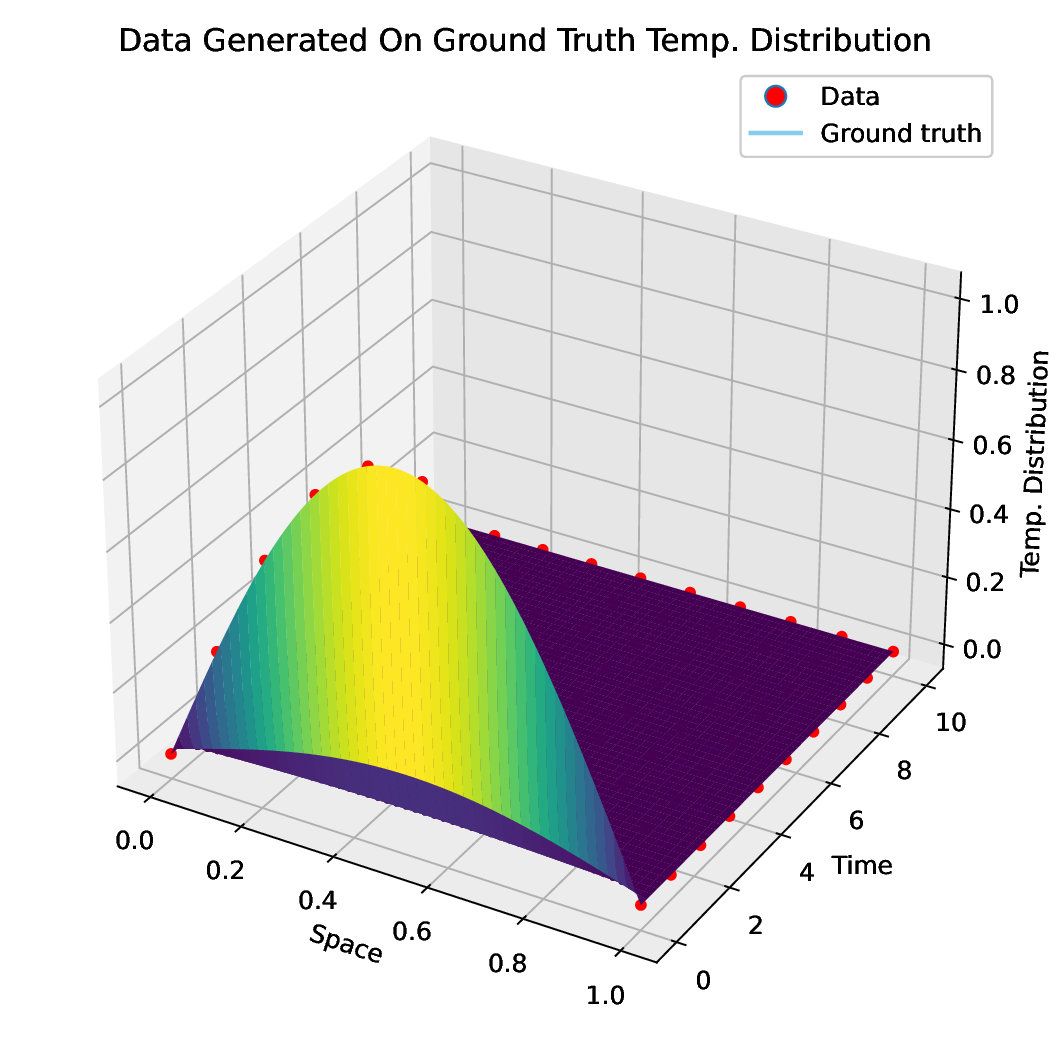}  
  \caption{Data with D = $1$}
  \label{fig:temp_space_1_3d}
\end{subfigure}
\caption{Temperature distribution over space and time for 3 different datasets with D values~=~$0.01, 0.1$, and $1$ respectively}
\label{fig:diff_d_data}
\end{figure}

In Figure \ref{fig:diff_d_data} the data that our network is being trained on is vastly different due to the varying thermal diffusivity value that we set.

The differences that Figure \ref{fig:diff_d_data} highlight are caused the changes in how quickly the temperature concavity is smoothed out. The higher thermal diffusivity values correlate to the data smoothing out much quicker compared to lower thermal diffusivity values.

Once we have determined our data, we can test out models with different thermal diffusivity values and examine how varying the rate of concave smoothening affects the accuracy of our models.

\begin{figure}[htbp]
\begin{subfigure}{.5\textwidth}
  \centering
  \includegraphics[width=\linewidth]{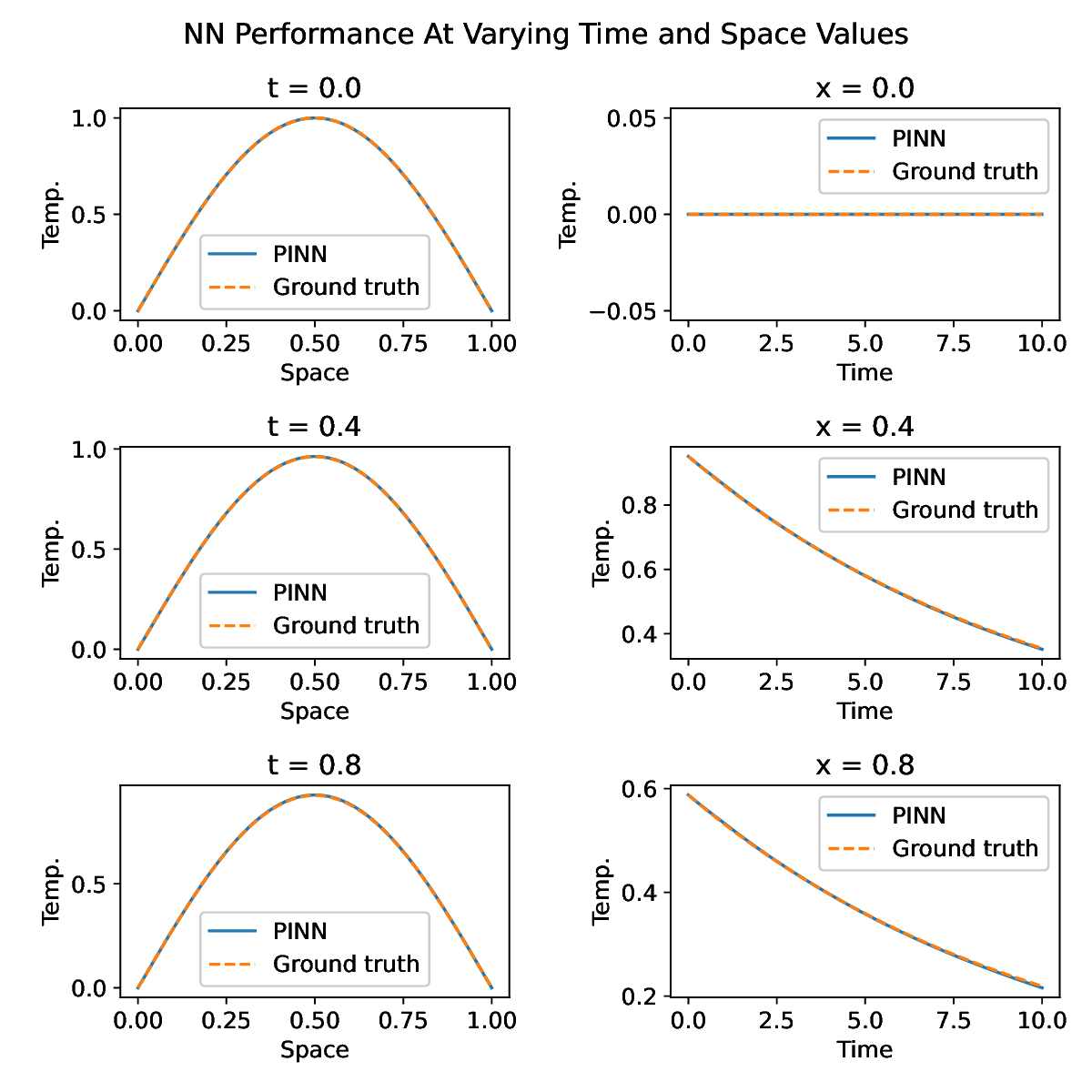}  
  \caption{Model with D = $0.01$}
  \label{fig:temp_space_0.01}
\end{subfigure}
\begin{subfigure}{.5\textwidth}
  \centering
  \includegraphics[width=\linewidth]{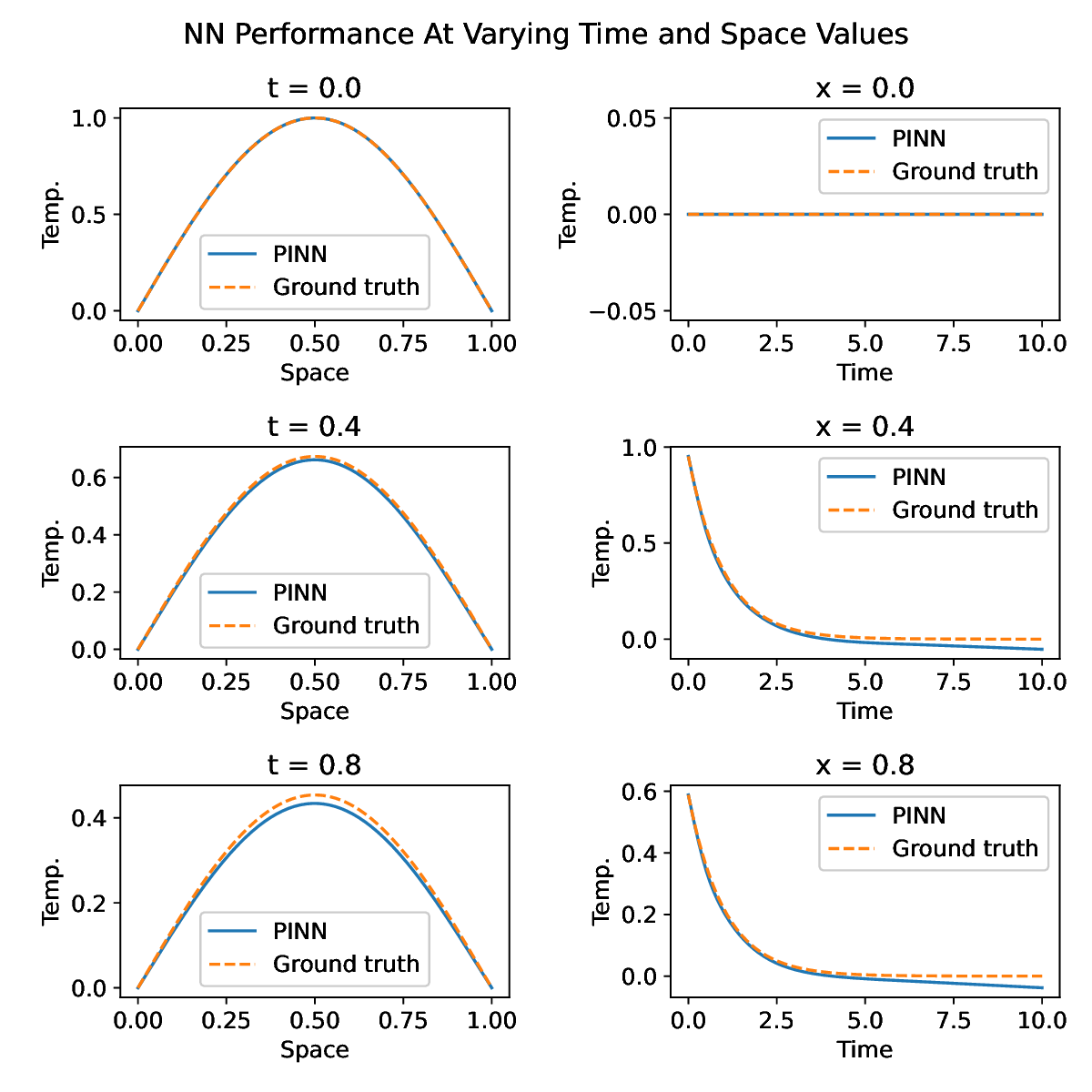}  
  \caption{Model with D = $0.1$}
  \label{fig:temp_space_0.1}
\end{subfigure}
\begin{subfigure}{.5\textwidth}
  \centering
  \includegraphics[width=\linewidth]{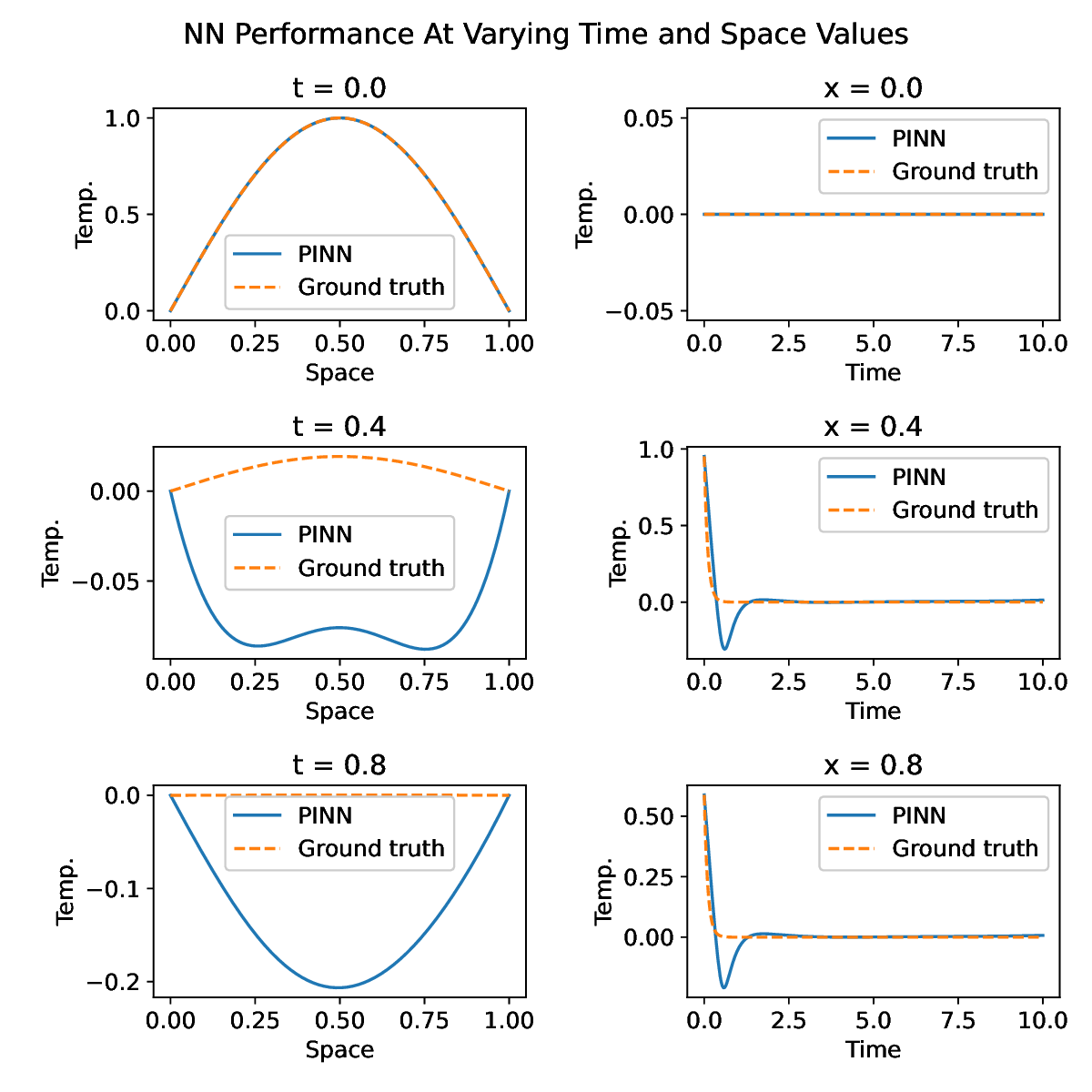}  
  \caption{Model with D = $1$}
  \label{fig:temp_space_1}
\end{subfigure}
\caption{Temperature distribution over space for 3 different models with D~=~$0.01, 0.1$, and $1$ respectively}
\label{fig:diff_d_val}
\end{figure}

In Figure \ref{fig:diff_d_val} we examine models with different thermal diffusivity values by looking at their temperature distribution over space.

Evidently, Figure \ref{fig:diff_d_val} shows that two models with thermal diffusivity values $0.01$ and $0.1$ often fit the model well. However, when we use the thermal diffusivity value of $1$ the neural network tends to stray from the ground truth.  This is due to the rapid change in our data with a higher thermal diffusivity value. With a value of $1$, the concavity smooths out quicker than models with a lower thermal diffusivity.

It is important to note that the range of the sub-figure with the thermal diffusivity value of $1$ is tiny relative to the other sub-figures. Therefore, although the model looks extremely inaccurate, the loss is still small.

However, the main reason the latter model performs so poorly is due to the grid density that the model trains on. If the data given changes rapidly, and the density of the grid is small, it becomes difficult for the next work to understand trends in the data.

In Table \ref{tab:density_values} we examine three different models trained on grids or datasets with three different density values corresponding to how many data points there are in each dataset. The relationship between the density value and how dense our grid is inversely proportional, as we create our dataset with the term $1/D_{density}$. We calculate the average MSE and total loss assuming that our thermal diffusivity value remains constant throughout all three models.

\begin{table}[htbp]
    \centering
    \begin{tabular}{c|c|c}
        Density Value & Avg MSE & Avg Total Loss \\
        \hline
        0.1 & $2.46*10^{-6}$ & $1.18*10^{-5}$ \\
        \hline
        0.05 & $2.34*10^{-6}$ & $1.42*10^{-5}$ \\
        \hline
        0.025 & $2.52*10^{-9}$ & $1.03*10^{-6}$ \\
    \end{tabular}
    \caption{Table with 3 different density values with average MSE and average Total Loss calculated. We assume that the thermal diffusivity is $0.1$. The lower the density the more number of points in our dataset as our dataset length is determined by the term $1/D_{density}$.}
    \label{tab:density_values}
\end{table}

As Table \ref{tab:density_values} illustrates, the denser our dataset, the better our model does. This intuitively makes sense, as a denser dataset would allow the model to more easily capture deeper trends in data with smaller increments.

\subsubsection{Implementing the Inverse Problem}
We retrained the same model after adding thermal diffusivity (D) as a parameter in the neural network. This small change in the training process allowed us to also solve the inverse problem, meaning we could determine the value for D that fits our given data most optimally.

Table \ref{tab:heat_inv_metrics} reflects some different metrics that provide insight into how the model performed. Both measurements of error (ground truth MSE and total loss) are low, and the D value after training was remarkably close to 0.1 from our exact solution. The average runtime of 72.627 also communicates the model is relatively efficient.

\begin{table}[htbp]
    \centering
    \begin{tabular}{c|c}
        Metric & Average \\
        \hline
        $[0, 1]$ Ground Truth MSE & $3.774*10^{-6}$ \\
        \hline
        Total Loss & $2.963*10^{-5}$ \\
        \hline
        D After Training & $0.0999$ \\
        \hline
        Time (sec.) & $72.627$
    \end{tabular}
    \caption{4 different metrics to describe our Heat Equation PINN's performance. Averages were calculated over 10 iterations.}
    \label{tab:heat_inv_metrics}
\end{table}

\begin{figure}[htp]
    \centering
    \makebox[\textwidth][c]{\includegraphics[width=0.8\textwidth]{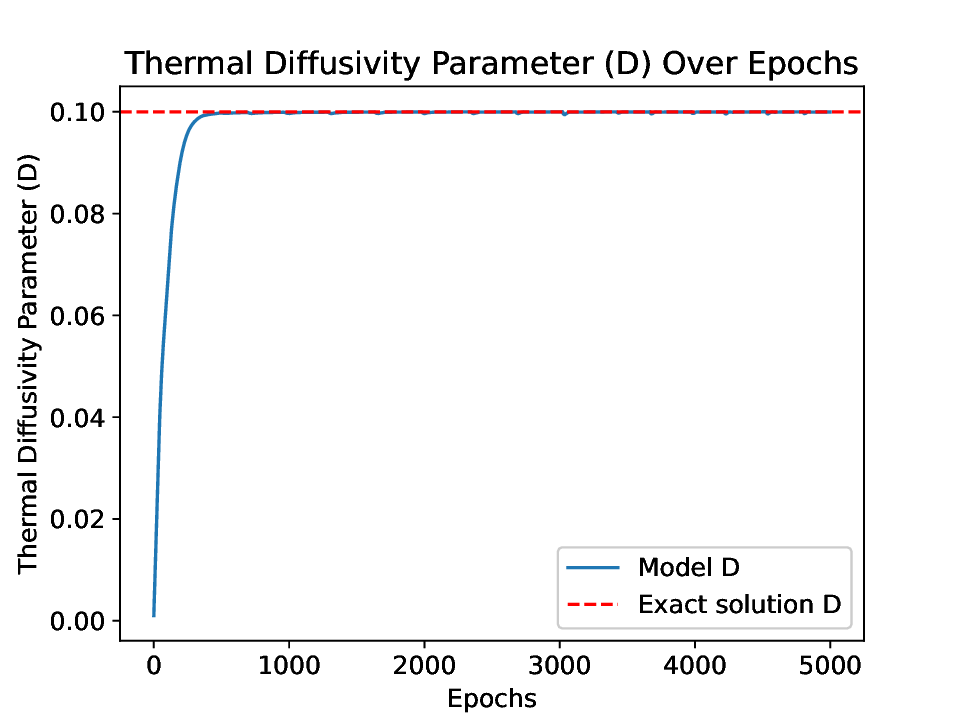}}
    \caption{Thermal Diffusivity at different Epochs during the training of the PINN}
    \label{fig:d_vals_epochs}
\end{figure}

We also took note of our thermal diffusivity value during the training process. Using our exact solution value of $0.1$ we noted the D value of our neural network at a specific epoch in time in Figure \ref{fig:d_vals_epochs}

Evidently, the thermal diffusivity value research the exact solution and stays there, which denotes our PINN being trained effectively.

\begin{figure}[htp]
    \centering
    \makebox[\textwidth][c]{\includegraphics[width=0.8\textwidth]{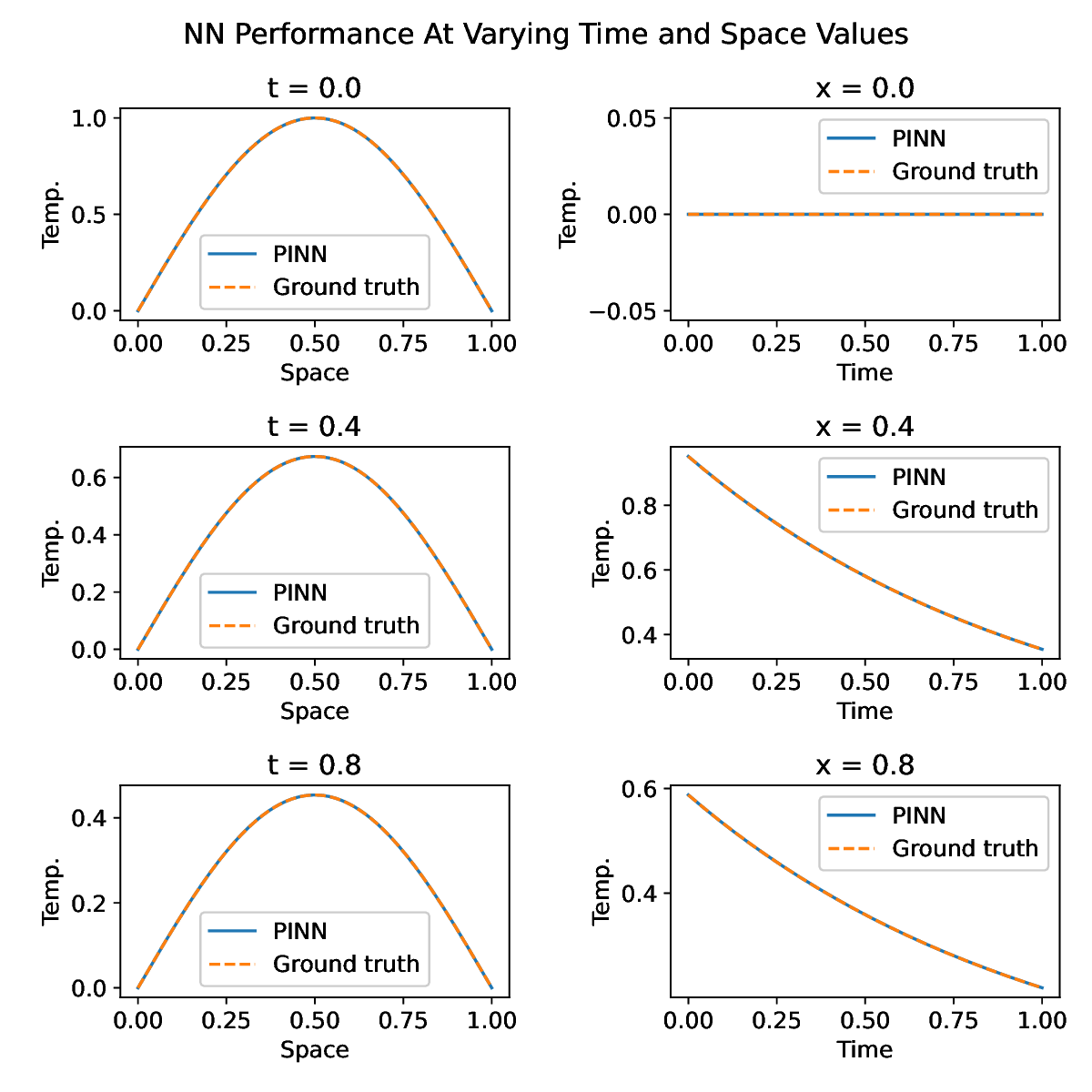}}
    \caption{PINN at different time and space values. Specifically when time and space is at $0.0, 0.4$, and $0.8$}
    \label{fig:pinn_interval}
\end{figure}

\begin{figure}[htp]
    \centering
    \makebox[\textwidth][c]{\includegraphics[width=0.8\textwidth]{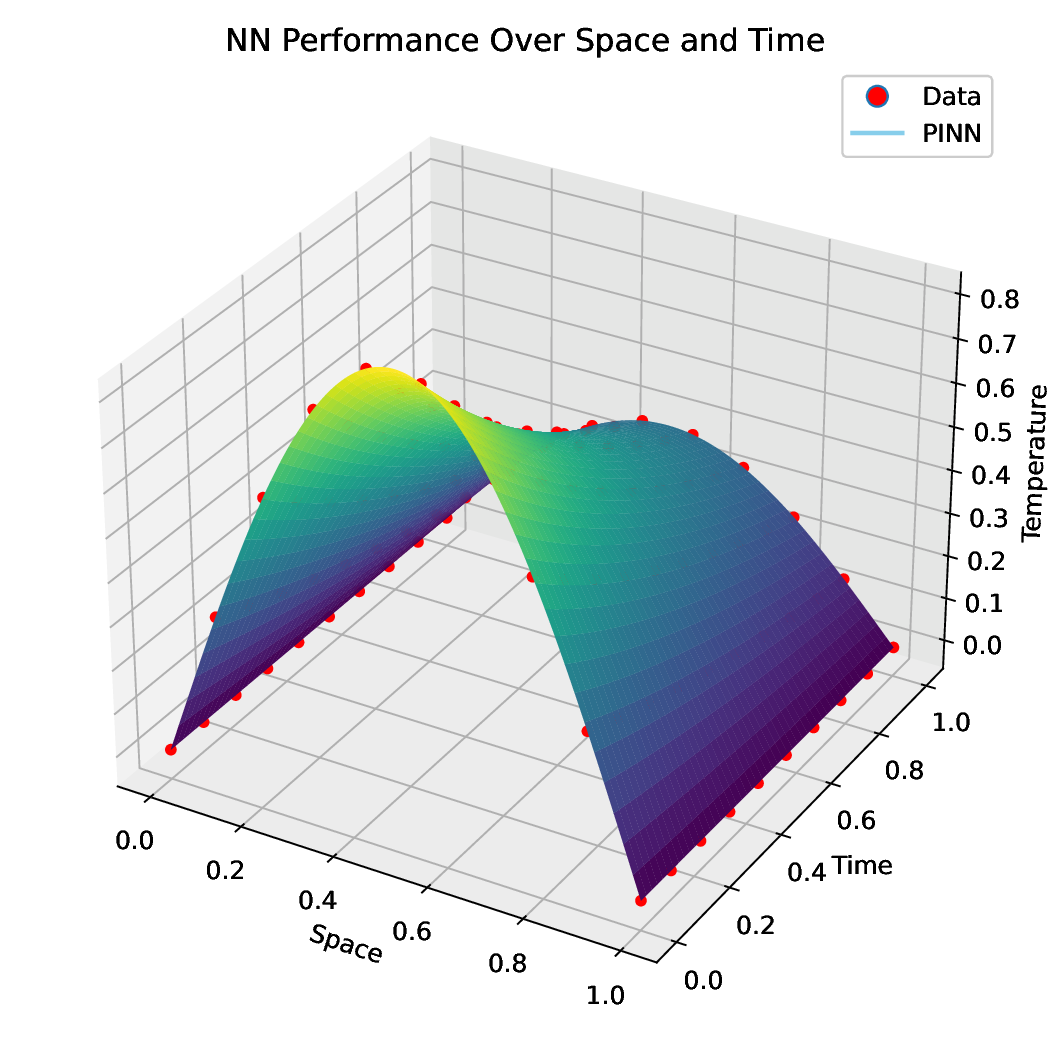}}
    \caption{PINN evaluated over the 3D grid with time, space, and the temperature distribution}
    \label{fig:pinn_3d}
\end{figure}

We also plotted the neural network at different time and space values and over a 3D grid as shown in Figure \ref{fig:pinn_interval} and Figure \ref{fig:pinn_3d}

\begin{figure}[htp]
    \centering
    \makebox[\textwidth][c]{\includegraphics[width=0.8\textwidth]{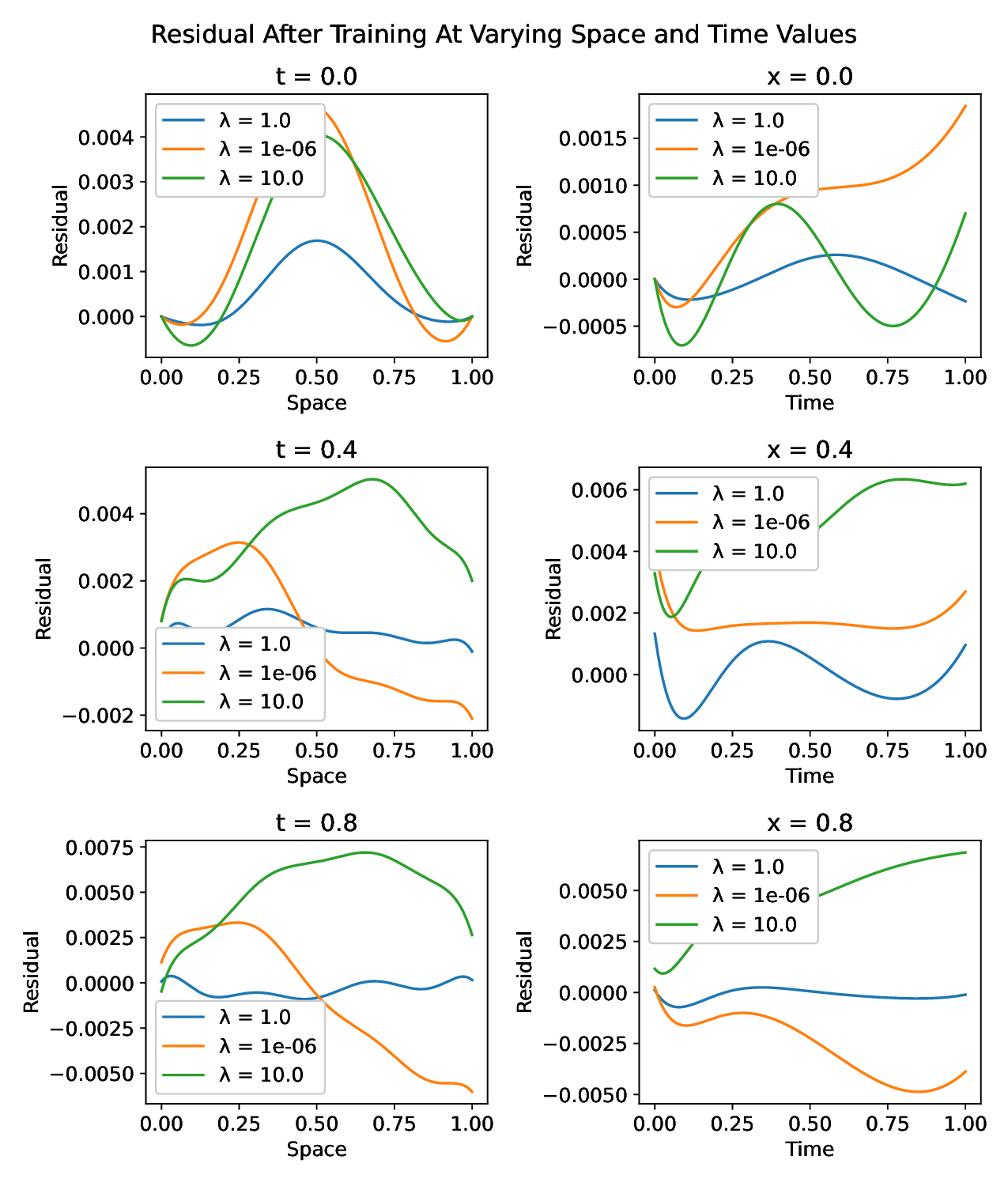}}
    \caption{The residual at different space and time values which model the same space and time values in Figure \ref{fig:pinn_interval}. We modeled the residual at different strengths to understand the impact it can have on our final product.}
    \label{fig:heat_inverse_res}
\end{figure}

Through the figures we concluded that the PINN has also been evaluated efficiently, and finally we plotted our residuals to understand how the residual changes our final product, which is reflected in Figure \ref{fig:heat_inverse_res}.

We concluded that when $\lambda = 1.0$ our model performs the best, and we also concluded that our PINN can effectively solve both the forward and inverse problems, making it a recommended technique for solving differential equations.

\subsubsection{FDM and PINN Comparison}\label{subsubsec:fdm_pinn_results}
We next explored a condition for $\Delta x$, $\Delta t$, and $D$ that may be required for a stable model. As described in Section \ref{subsec:fdm}, we investigated this condition for the FDM and PINN qualitatively and quantitatively.

In one instance, we set $\Delta x=\Delta t=0.05$ and the final time value to 10. Then, we varied values of $D$. Table \ref{tab:heat_dxdt_0.05_t_10} compares the PINN and FDM performance. 0.025 was the largest value for D that would not break the CFL condition, and we saw in the table that when a greater value $D=0.045$ was tested, the FDM MSE increased dramatically. However, we did not see the same pattern in the PINN. While the MSE increased with $D$, values that broke the CFL condition did not significantly affect the PINN. 

\begin{table}[htbp]
    \centering
    \begin{tabular}{c|c|c}
        D & PINN Data Point MSE & FDM MSE \\
        \hline
        $0.005$ & $5.134*10^{-8}$ & $1.277*10^{-8}$\\
        \hline
        $0.025$ & $5.613*10^{-8}$ & $7.127*10^{-7}$\\
        \hline
        $0.045$ & $1.558*10^{-7}$ & $3.520*10^{128}$\\
    \end{tabular}
    \caption{Comparison of PINN and FDM approaches at varying $D$ when $\Delta x=\Delta t=0.05$ and maximum time was 10.}
    \label{tab:heat_dxdt_0.05_t_10}
\end{table}

\begin{figure}[htbp]
    \begin{subfigure}{.55\linewidth}
      \centering
      \includegraphics[width=\linewidth]{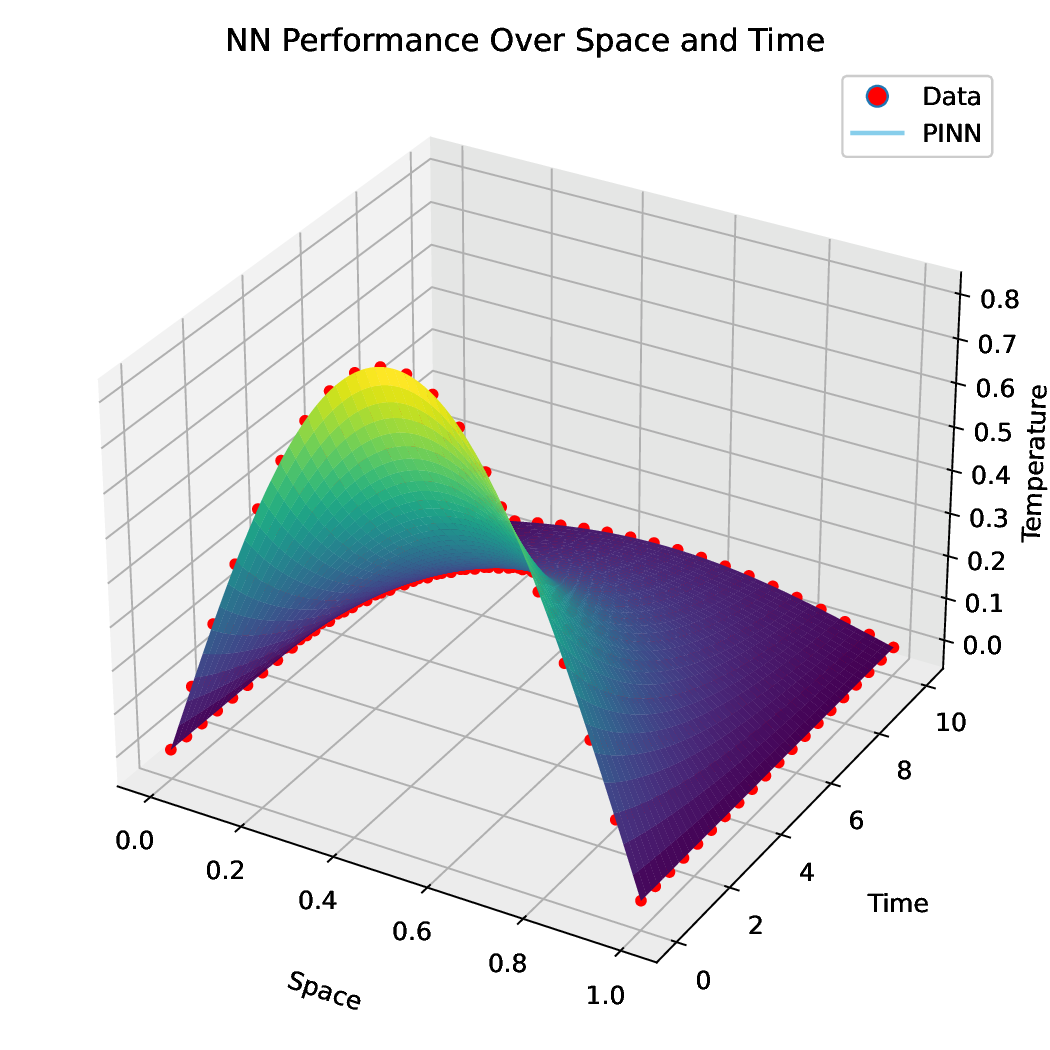}
      \caption{PINN Result}
      \label{fig:pinn_dxdt_0.05_D_0.025}
    \end{subfigure}
    \begin{subfigure}{.6\linewidth}
      \centering
      \includegraphics[width=\linewidth]{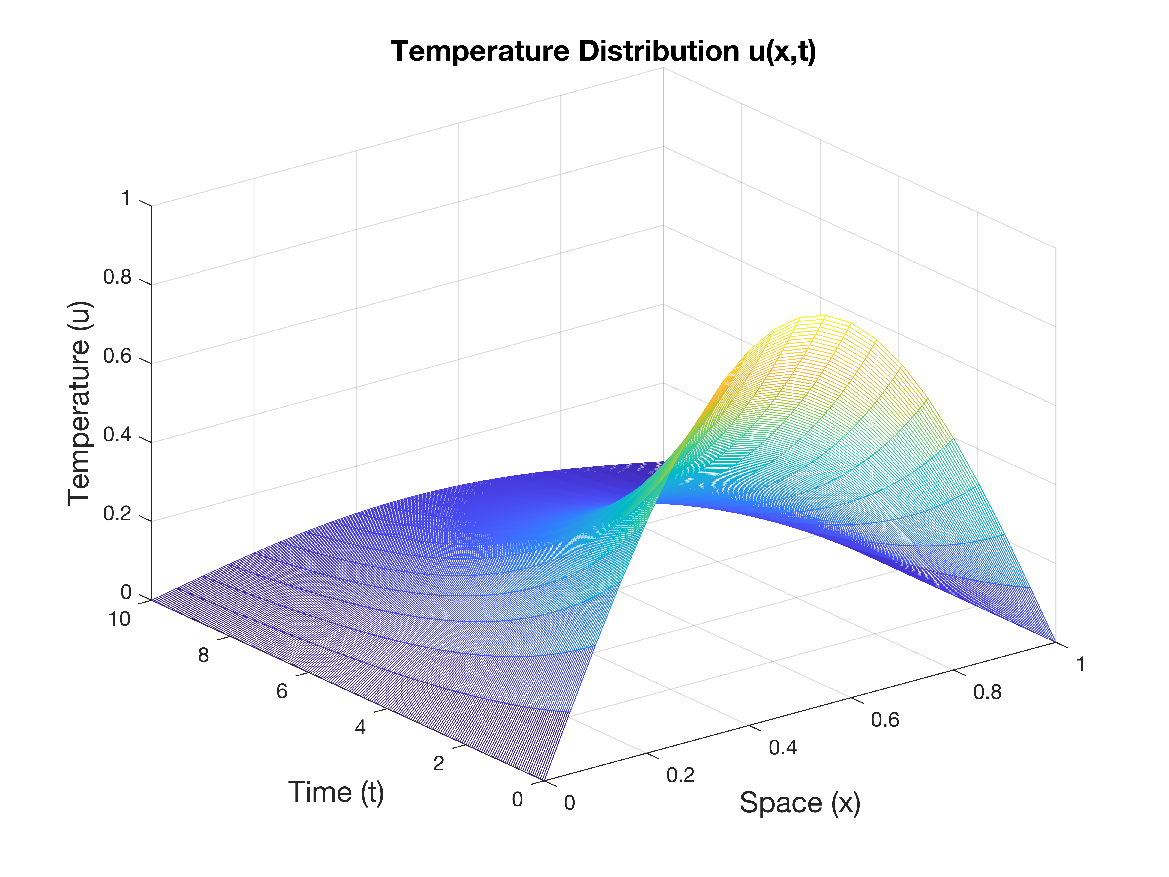}  
      \caption{FDM Result}
      \label{fig:fdm_dxdt_0.05_D_0.025}
    \end{subfigure}
\caption{Visual comparison of PINN and FDM approaches at $D=0.025$ when $\Delta x=\Delta t=0.05$ and maximum time was 10. The CFL condition was satisfied at this configuration.}
\label{fig:heat_dxdt_0.05_t_10_D_0.025}
\end{figure}

We also analyzed the models visually using 3D plots. Figure \ref{fig:heat_dxdt_0.05_t_10_D_0.025} shows us their performance when the CFL condition is still satisfied. Both approaches yield stable results. Table \ref{tab:heat_dxdt_0.05_t_10} communicates their accuracy.

\begin{figure}[htbp]
    \begin{subfigure}{.55\linewidth}
      \centering
      \includegraphics[width=\linewidth]{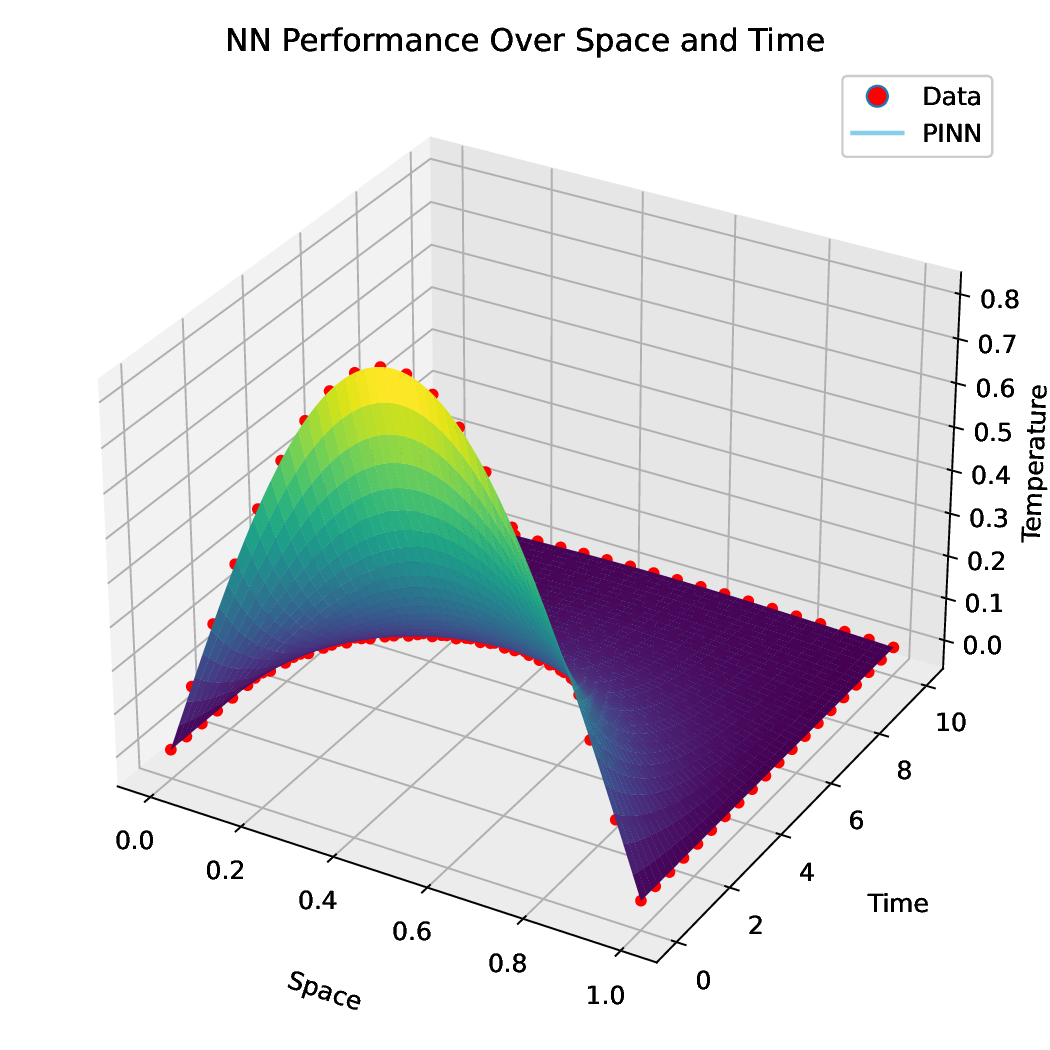}
      \caption{PINN Result}
      \label{fig:pinn_dxdt_0.05_D_0.045}
    \end{subfigure}
    \begin{subfigure}{.6\linewidth}

      \centering
      \includegraphics[width=\linewidth]{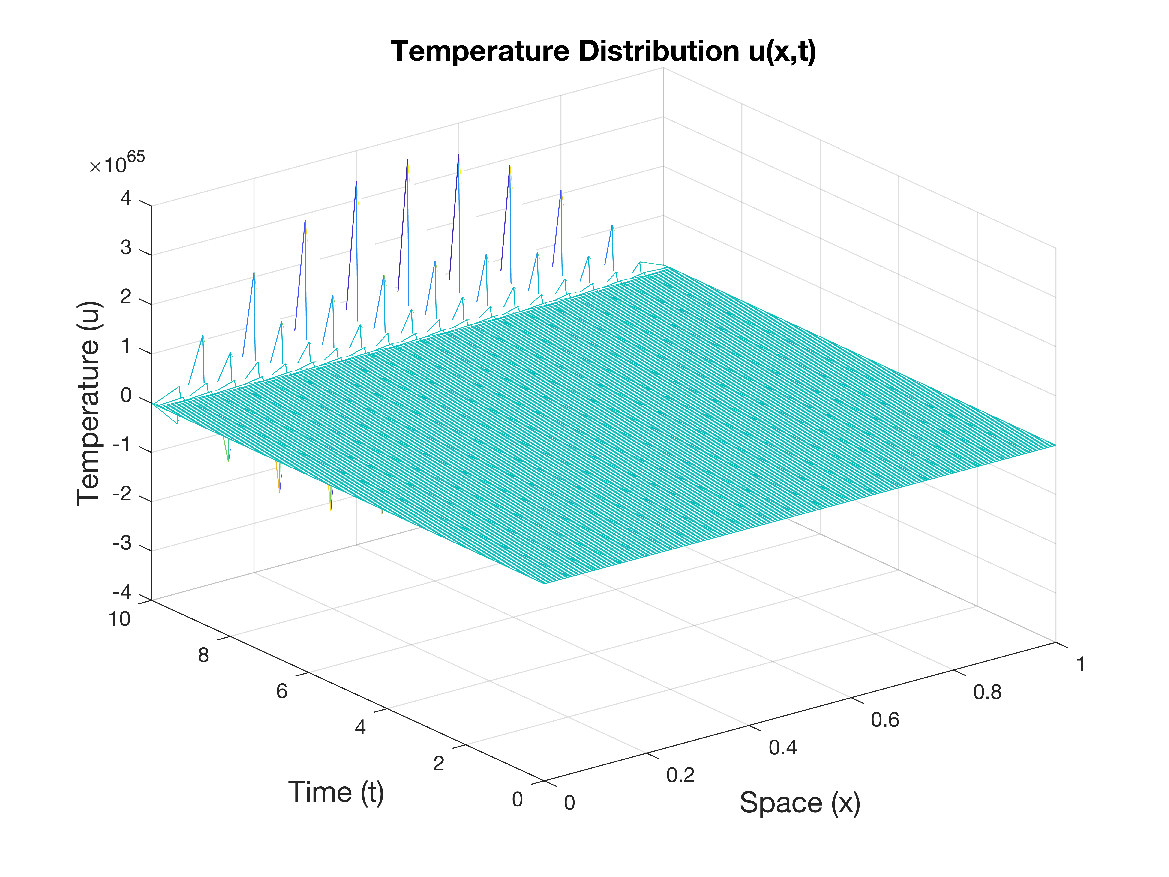}  
      \caption{FDM Result}
      \label{fig:fdm_dxdt_0.05_D_0.045}
    \end{subfigure}
\caption{Visual comparison of PINN and FDM approaches at $D=0.045$ when $\Delta x=\Delta t=0.05$ and total time was 10. The CFL condition was broken.}
\label{fig:heat_dxdt_0.05_t_10_D_0.045}
\end{figure}

We then looked at plots that broke the CFL condition, such as when $D=0.045$. Figure \ref{fig:heat_dxdt_0.05_t_10_D_0.045} displays the difference between the methods. As shown in Figure \ref{fig:fdm_dxdt_0.05_D_0.045}, the model blew up and did not retain the shape of the exact solution. However, Figure \ref{fig:pinn_dxdt_0.05_D_0.045} showed us that the PINN did not experience this impact.

\begin{figure}[htp]
    \centering
    \makebox[\textwidth][c]{\includegraphics[width=0.7\textwidth]{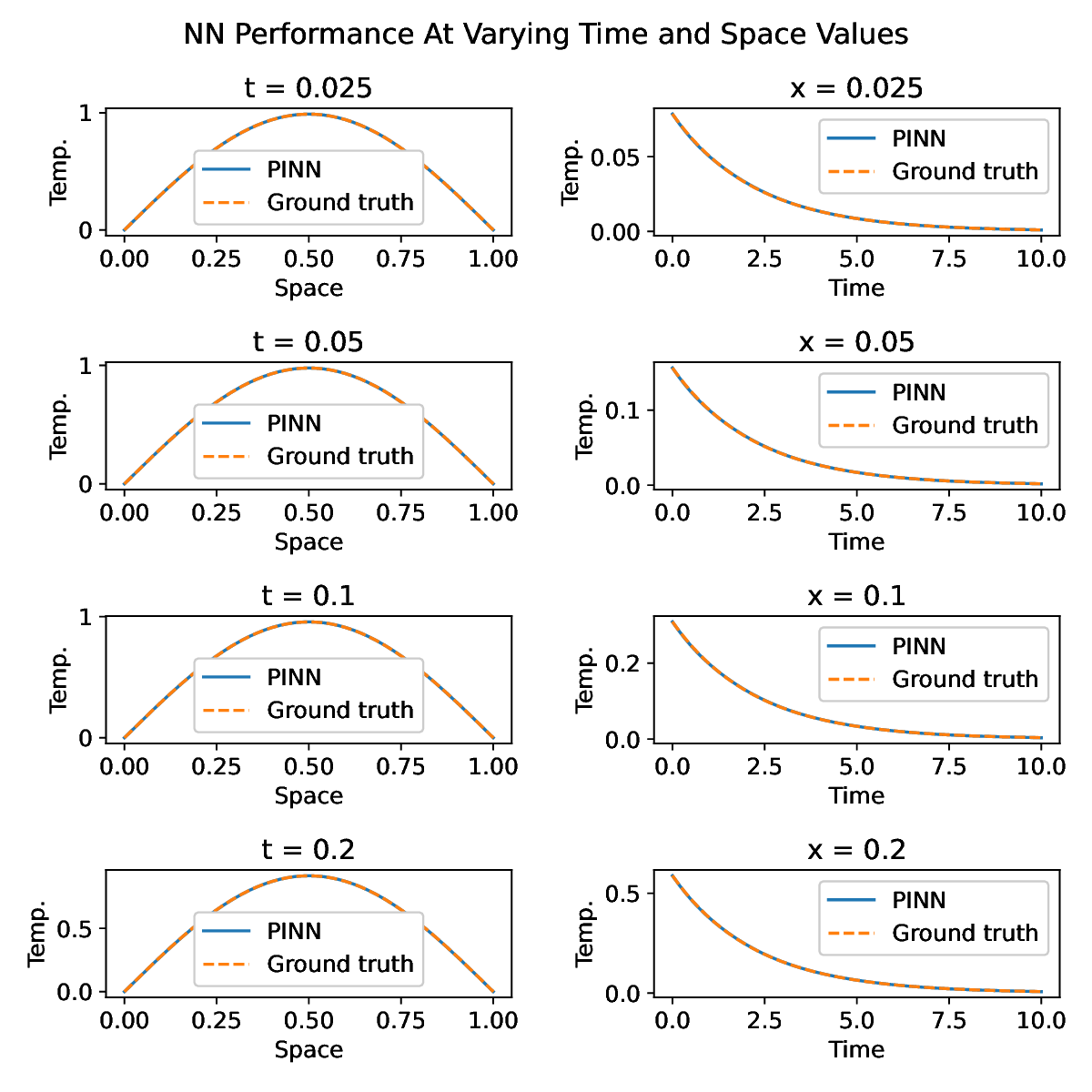}}
    \caption{Performance of PINN at fixed time and space values when $D=0.045$, $\Delta x=\Delta t=0.05$, and total time was 10. The CFL condition was broken.}
    \label{fig:heat_pinn_slices_dxdt_0.05_D_0.045}
\end{figure}

Experimenting with both approaches showed us that the PINN does not follow the same behavior as the FDM concerning stability and accuracy past the CFL condition. A closer look at the PINN in Figure \ref{fig:heat_pinn_slices_dxdt_0.05_D_0.045} further communicates its accuracy. The model's result is hard to distinguish from the ground truth line despite the broken condition. Although we know that the MSE increased with $D$, this figure tells us that the PINN performance was still strong.

We concluded that the PINN performed differently than the FDM when evaluated near the CFL condition. Due to the way the solution is calculated, the PINN likely does not depend as much on the relationship between grid density and thermal diffusivity.

We performed the same analysis at different grid densities, maximum time values, and $D$ values. Refer to Appendix \ref{subsec:fdm_pinn} for details regarding those results.

\section{Discussion}

Leveraging prior knowledge in neural networks has widespread applications across fields such as physics, engineering, and medical sciences. Using differential equations to implement prior information, PINNs perform accurately and reliably on sparse, high-dimensional data. These neural networks are equally helpful when fitting noisy data because they add the component models that often struggle to learn: the larger trend. Prior information closes this important gap between data and the exact solution, making it an invaluable addition to neural networks.

A numerical solution method that solves PDEs is a finite difference method (FDM) \cite{ammari_lecture_nodate}, and a PINN is a promising alternative to it. In the scientific community, PINNs are considered important multi-task frameworks because the model can solve a PDE like an FDM in addition to fitting given data \cite{cuomo_scientific_2022}. Existing literature such as \cite{zhang_personalized_2024} establishes their applications to complex differential equations like a biophysical PDE model of the glioblastoma tumor.

Given the existing appreciation for PINNs in achieving the tasks of fitting data and solving differential equations, we wanted to emphasize their versatility across different types of problems. The linear and quadratic models serve as the simplest examples of PINNs, showing the similarity between known regularization techniques \cite{tian_comprehensive_2022} and implementing a residual. Our work draws on this similar relationship to portray the application of PINNs to synthetic data of varying complexity. In the linear and quadratic examples, the differential equations consisted of one derivative. The most complex example, the Heat Equation PINN, consists of two partial derivatives and communicates that the model easily extends to more involved differential equations. Furthermore, the Heat Equation PINN reinforces the additional ability of these networks to solve the inverse problem and estimate equation parameters. 

Making use of more complicated examples from literature, our systematic approach to establishing the computational power of the PINN approach makes our work interesting.

\section{Conclusion}

In our work, we examined the effectiveness of PINNs in solving the forward problem across different examples, starting by fitting linear and quadratic data points and later working with data that models the one-dimensional heat equation differential equation. By both fitting data successfully and solving the given differential equation, our PINNs proved their capability to complete both tasks simultaneously.

We additionally solved the inverse problem for the Heat Equation PINN,  This final work reinforced an additional benefit of these networks; with no knowledge about the equation parameter $D$, the slightly-modified PINN could still successfully fit synthetic data points. 

In both simpler and more complex scenarios, our exploration serves as a deep dive into important components of PINNs and neural networks such as the strength of the residual $\lambda$ as well as layer/neuron configurations. We establish that while higher $\lambda$ values can improve model performance, there is a value too high, and an important balance between the data and residual losses must be achieved. Understanding the behavior of $\lambda$ is essential to producing effective PINNs. Similarly, we displayed the impact of layer/neuron configurations by testing different models and calculating their losses. Our method of determining the optimal network architecture, while not perfect, communicates the importance of catering the layer/neuron configuration to the problem at hand; as we increased the complexity of the data from linear to quadratic to that of the one-dimensional heat equation, the models required more layers and neurons.

Another important result of our research, as shown in Section \ref{subsubsec:fdm_pinn_results}, is that our Heat Equation PINN does not have the same requirements as an FDM concerning the CFL condition. While the FDM quickly became unstable as the CFL condition was broken, the PINN displayed slight inaccuracies while generally adhering to the input data. 

Looking forward, given the results from the FDM and PINN comparison, we would like to find an equivalent of the CFL condition for our Heat Equation PINN for accuracy and stability. If such a condition does not exist, we are interested in why that is.

Next, we hope to extend our work with the applications of PINNs, developing a Glioblastoma Infiltration PINN on MRI scans similar to \cite{zhang_personalized_2024}. By incorporating biological knowledge into our networks, we want to predict trends with little experimental data as often is the case with real patient cases. 

\section{Acknowledgments}

We would like to thank our principal investigator Dr. John Lowengrub and teaching assistant Dr. Ray Zirui Zhang at the Department of Mathematics at the University of California, Irvine (UCI) for their guidance and support throughout the research process. We also acknowledge Deborah Tonne, Dr. Alessandra Pantano, and Dr. German Enciso at UCI's Math ExpLR program for the opportunity to participate in mathematical research. 

\section{Appendix}

\subsection{Details on Linear PINN Architecture}\label{subsec:optimallinear}

We used the method outlined in Section \ref{subsec:nn} to determine a satisfactory number of layers and neurons for the PINN. The first part of this method is shown in Table \ref{tab:linear_4l_gt}. It entails 5 configurations with 4 layers and varying neurons.

\begin{table}[htbp]
    \centering
    \begin{tabular}{c|c|c|c}
        Layers & Neurons & Avg. Ground Truth MSE & Standard Deviation\\
         \hline
        4 & 5 & $1.502*10^{-2}$ & $1.111*10^{-2}$\\
        \hline
        4 & 10 & $1.404*10^{-2}$ & $1.125*10^{-2}$\\
        \hline
        4 & 20 & $\boldsymbol{1.375*10^{-2}}$ & $1.112*10^{-2}$ \\
        \hline
        4 & 30 & $1.412*10^{-2}$ & $1.096*10^{-2}$ \\
        \hline
        4 & 40 & 
        $\boldsymbol{1.368*10^{-2}}$ & $9.809*10^{-3}$\\
    \end{tabular}
    \caption{Average standard deviation and MSE calculated with ground truth function over $[-1, 2]$. Error is calculated for models with 4 fixed layers and varying neurons.}
    \label{tab:linear_4l_gt}
\end{table}

Another helpful way to analyze error is using the total loss, which contains data and residual loss from training. We would rely on this metric in a scenario where we do not have the ground truth function. However, the ground truth function is more accurate for error calculation because it includes more information about the data trend(s). The average total loss is shown in Table \ref{tab:linear_4l_total}.

\begin{table}[htbp]
    \centering
    \begin{tabular}{c|c|c|c}
        Layers & Neurons & Avg. Total Loss & Standard Deviation\\
         \hline
        4 & 5 & $9.786*10^{-3}$ & $5.870*10^{-3}$\\
        \hline
        4 & 10 & $9.773*10^{-3}$ & $4.981*10^{-3}$\\
        \hline
        4 & 20 & $9.771*10^{-3}$ & $4.982*10^{-3}$ \\
        \hline
        4 & 30 & $\boldsymbol{9.769*10^{-3}}$ & $4.982*10^{-3}$ \\
        \hline
        4 & 40 & $9.809*10^{-3}$ & $4.936*10^{-3}$\\
    \end{tabular}
    \caption{Average total loss and standard deviation for models with 4 fixed layers and varying neurons.}
    \label{tab:linear_4l_total}
\end{table}

Table \ref{tab:linear_4l_gt} tells us that a model with 4 layers and 40 neurons performs better than the other models. That configuration leads to a ground truth MSE of $1.368*10^{-2}$. We could proceed to manipulate the layers and fix 40 neurons, but it is important to note that the MSE for 20 neurons is $1.375*10^{-2}$. The difference in MSEs for those configurations is marginal. To decide between the two, we considered runtime and our knowledge about neural networks. In one run of these 5 models, the training times for 20 and 40 neurons were 19.843 seconds and 20.239 seconds, respectively. We also know that simpler models can overfit with excessive neurons. These two factors led us to choose 20 neurons as the optimal number for 4 layers.

If we instead used the average total loss as shown in Table \ref{tab:linear_4l_total}, the optimal number of neurons would be 30. While this is different than the results from Table \ref{tab:linear_4l_gt}, we see that the differences in average total loss are not that large. For our purposes, choosing 30 neurons would not have been sub-optimal because our goal is a configuration that fits the data relatively well. We thus followed our initial method and chose 20 neurons.

As shown in Table \ref{tab:linear_20n_gt}, we next manipulated the number of layers while keeping 20 neurons fixed. 

\begin{table}[htbp]
    \centering
    \begin{tabular}{c|c|c|c}
        Layers & Neurons & Avg. Ground Truth MSE & Standard Deviation\\
         \hline
        2 & 20 & $6.023*10^{-3}$ & $6.778*10^{-3}$\\
        \hline
        3 & 20 & $\boldsymbol{6.020*10^{-3}}$ & $6.526*10^{-3}$\\
        \hline
        4 & 20 & $6.578*10^{-3}$ & $6.711*10^{-3}$ \\
        \hline
        5 & 20 & $6.907*10^{-3}$ & $6.706*10^{-3}$ \\
        \hline
        6 & 20 & $8.648*10^{-3}$ & $6.037*10^{-3}$\\
    \end{tabular}
    \caption{Average standard deviation and MSE calculated with ground truth function over $[-1, 2]$. Error is calculated for models with 20 fixed neurons and varying layers.}
    \label{tab:linear_20n_gt}
\end{table}

Table \ref{tab:linear_20n_gt} highlights that within our configurations for 20 neurons, 3 layers resulted in the lowest average ground truth MSE. The difference between 2 and 3 layers was not drastic, so either could have been chosen. The loss for both is small, so we opted for 3 layers. 

Similar to our analysis of varying neurons, we also looked at the average total loss in Table \ref{tab:linear_20n_total} to see how those results compared. 

\begin{table}[htbp]
    \centering
    \begin{tabular}{c|c|c|c}
        Layers & Neurons & Avg. Total Loss & Standard Deviation\\
         \hline
        2 & 20 & $1.19475*10^{-2}$ & $5.238*10^{-3}$\\
        \hline
        3 & 20 & $1.19225*10^{-2}$ & $5.247*10^{-3}$\\
        \hline
        4 & 20 & $1.19228*10^{-2}$ & $5.248*10^{-3}$ \\
        \hline
        5 & 20 & $\boldsymbol{1.19223*10^{-2}}$ & $5.246*10^{-3}$ \\
        \hline
        6 & 20 & $\boldsymbol{1.19223*10^{-2}}$ & $5.247*10^{-3}$\\
    \end{tabular}
    \caption{Average total loss and standard deviation for models with 20 fixed neurons and varying layers.}
    \label{tab:linear_20n_total}
\end{table}

Between the different numbers of layers for 20 neurons, there was minimal variation between the average total loss values. If our criterion for optimal layers or neurons was average total loss, 5 or 6 layers would be optimal. However, the results for 3, 4, 5, and 6 layers were all similar and therefore 3 layers was still an informed choice. Also, as mentioned, ground truth MSE over $[-1, 2]$ is a more reliable measure than average loss. 

We made our decisions about optimal layers and neurons based on the ground truth MSE over the interval $[-1, 2]$. This includes data inside and outside of the training range. However, in an individual run of all 10 layer and neuron configurations, we also examined ground truth MSE exclusively within the training range $[0, 1]$. Our goal was to see if minimizing the $[0, 1]$ ground truth MSE would lead to a different outcome. We regardless chose the larger interval $[-1, 2]$ because we wanted our model to generalize well to new data. Analyzing the error on data outside the training range helped us achieve that goal. Table \ref{tab:linear_mse_int} nevertheless shows the alternate $[0, 1]$ ground truth perspective.

\begin{table}[htbp]
    \centering
    \begin{tabular}{c|c|c|c}
        Layers & Neurons & $[-1, 2]$ Ground Truth MSE & $[0, 1]$ Ground Truth MSE\\
         \hline
        4 & 5 & $2.256*10^{-2}$ & $2.285*10^{-3}$\\
        \hline
        4 & 10 & $1.981*10^{-2}$ & $2.236*10^{-3}$\\
        \hline
        4 & 20 & $\boldsymbol{1.898*10^{-2}}$ & $2.226*10^{-3}$ \\
        \hline
        4 & 30 & $2.235*10^{-2}$ & $2.247*10^{-3}$ \\
        \hline
        4 & 40 & $2.014*10^{-2}$ & $\boldsymbol{2.115*10^{-3}}$\\
        \hline
        2 & 20 & $\boldsymbol{6.574*10^{-3}}$ & $1.6623*10^{-3}$\\
        \hline
        3 & 20 & $6.830*10^{-3}$ & $\boldsymbol{1.6418*10^{-3}}$\\
        \hline
        4 & 20 & $6.976*10^{-3}$ & $1.6423*10^{-3}$ \\
        \hline
        5 & 20 & $7.222*10^{-3}$ & $1.6420*10^{-3}$ \\
        \hline
        6 & 20 & $7.197*10^{-3}$ & $1.6549*10^{-3}$\\
    \end{tabular}
    \caption{Ground truth MSE for one iteration calculated in two ways: throughout the interval $[-1, 2]$ and exclusively within the training interval $[0, 1]$. The first method was used for determining optimal layers and neurons. This table only shows one run of the models, so its results for $[-1, 2]$ do not match those of Table \ref{tab:linear_4l_gt} and \ref{tab:linear_20n_gt} exactly. }
    \label{tab:linear_mse_int}
\end{table}

As shown in Table \ref{tab:linear_mse_int}, we can measure ground truth MSE in two ways and should be aware of how our choice impacts the decided layer and neuron configuration. In this run, both the optimal number of neurons and layers varied based on the interval. According to $[-1, 2]$ ground truth MSE, 20 neurons is optimal, but according to $[0, 1]$ ground truth MSE, 40 neurons is optimal. When 20 neurons was held constant, 2 layers was optimal according to $[-1, 2]$ but 3 layers was optimal according to $[0, 1]$. From this information we understood that our criterion for deciding layers and neurons can impact the final decision. Despite this limitation, the MSE values are similar and performance may not vary significantly.

We also find it important to note that between Table \ref{tab:linear_4l_gt} and \ref{tab:linear_20n_gt}, there is a discrepancy in average ground truth MSE for 4 layers and 20 neurons. The reason for the difference is that each run was random and the averages were calculated over 6-17 iterations; the variation is due to instances of uncertainty between two values after 5 iterations. When we had doubt, we ran more iterations to gain additional context. The standard deviation in those tables explain that the data was separately calculated.

Our final optimal network architecture for the Linear PINN was 3 layers and 20 neurons.

\subsection{Details on Quadratic PINN Architecture}\label{subsec:optimalquad}

We used the same method outlined in Section \ref{subsec:nn} as before to determine a satisfactory number of layers and neurons for the PINN. The first part of this method is shown in Table \ref{tab:quad_4layers}. It entails 5 configurations with 4 layers and varying neurons.

    \begin{table}[htbp]
        \centering
        \begin{tabular}{c|c|c|c}
        Layers & Neurons & Avg. Ground Truth MSE & Standard Deviation\\
         \hline
        4 & 5 & $5.868 *10^{-1}$ & $4.199*10^{-1}$\\
        \hline
        4 & 10 & $2.820*10^{-1}$ & $1.347*10^{-1}$\\
        \hline
        4 & 20 & $1.493*10^{-1}$ & $4.310*10^{-2}$ \\
        \hline
        4 & 30 & $\boldsymbol{1.011*10^{-1}}$ & $5.471*10^{-2}$ \\
        \hline
        4 & 40 & $1.535*10^{-1}$ & $7.615*10^{-2}$\\
        \end{tabular}
        \caption{Average standard deviation and MSE calculated with ground truth function over $[-1, 2]$. Error is calculated for models with 4 fixed layers and varying neurons.}
        \label{tab:quad_4layers}

    \end{table}

As shown in Table \ref{tab:quad_4layers}, when holding layers constant at 4, we can see that 30 neurons yields the best result, with the average ground truth MSE at $1.011*10^{-1}$. Unlike our linear model, the difference between 30 neurons and any other configuration is fairly vast. The standard deviation is just an indicator of variance between the points, as the model can fluctuate due to data and general variability within the model.

We also calculated the average total loss, which calculates the loss using the data instead of the ground truth function. This average loss is shown in Table \ref{tab:quad_4layers_total}.

    \begin{table}[htbp]
        \centering
        \begin{tabular}{c|c|c|c}
        Layers & Neurons & Avg. Total Loss & Standard Deviation\\
         \hline
        4 & 5 & $1.021 *10^{-2}$ & $2.739*10^{-3}$\\
        \hline
        4 & 10 & $6.477*10^{-3}$ & $3.293*10^{-3}$\\
        \hline
        4 & 20 & $5.554*10^{-3}$ & $2.509*10^{-3}$ \\
        \hline
        4 & 30 & $\boldsymbol{5.397*10^{-3}}$ & $2.666*10^{-3}$ \\
        \hline
        4 & 40 & $5.580*10^{-3}$ & $2.926*10^{-3}$\\
        \end{tabular}
        \caption{Average Total Loss results for models with 30 fixed neurons and varying layers. Total loss is calculated with the aggregate of the data loss and residual loss.}
        \label{tab:quad_4layers_total}

    \end{table}

As shown in Table \ref{tab:quad_4layers_total}, when holding layers constant at 4 and looking at average total loss, 30 neurons yield the best result with a loss of $5.397 * 10^{-3}$. 4 layers and 30 neurons fits our criteria, with the lowest average total loss and a comparable standard deviation to 40 neurons, which makes this neuron count optimal for our model. 

Comparing Table \ref{tab:quad_4layers} and \ref{tab:quad_4layers_total} it is evident that 30 neurons yields the best result on the ground truth line and on the data points that we generated.

In Table \ref{tab:quad_30neurons} we kept the number of neurons constant at 30 and changed the number of layers in our model to find the optimal layer count.

        \begin{table}[htbp]
        \centering
        \begin{tabular}{c|c|c|c}
        Layers & Neurons & Avg. Ground Truth MSE & Standard Deviation\\
         \hline
        2 & 30 & $7.199*10^{-1}$ & $2.821*10^{-1}$\\
        \hline
        3 & 30 & $1.740*10^{-1}$ & $6.274*10^{-2}$\\
        \hline
        4 & 30 & $1.580*10^{-1}$ & $9.549*10^{-2}$ \\
        \hline
        5 & 30 & $\boldsymbol{9.040*10^{-2}}$ & $5.189*10^{-2}$ \\
        \hline
        6 & 30 & $1.022*10^{-1}$ & $5.734*10^{-2}$\\
        \end{tabular}
        \caption{Average MSE results calculated over $[-1, 2]$ and standard deviation for models with 30 fixed neurons and varying layers.}
        \label{tab:quad_30neurons}
    \end{table}

In Table \ref{tab:quad_30neurons}, and we see that 5 layers would be the most optimal, as it has the lowest average ground truth MSE and the other lowest standard deviation at $9.040*10^{-2}$ and $5.189 * 10^{-2}$ respectively. Currently, our optimal layer count is 5 layers. As a result, we look at average total loss to determine the optimal network architecture. 

In Table \ref{tab:quad_30neurons_total} we will examine the average total loss and standard deviation while keeping the neuron count constant at 30 and varying and layer count.

    \begin{table}[htbp]
        \centering
        \begin{tabular}{c|c|c|c}
        Layers & Neurons & Avg. Total Loss & Standard Deviation\\
         \hline
        2 & 30 & $1.659*10^{-2}$ & $6.030*10^{-3}$\\
        \hline
        3 & 30 & $6.748*10^{-3}$ & $4.236*10^{-3}$\\
        \hline
        4 & 30 & $7.334*10^{-3}$ & $4.073*10^{-3}$ \\
        \hline
        5 & 30 & $\boldsymbol{6.547*10^{-3}}$ & $4.308*10^{-3}$ \\
        \hline
        6 & 30 & $6.570*10^{-3}$ & $4.299*10^{-3}$\\
        \end{tabular}
        \caption{Average Total Loss results for models with 30 fixed neurons and varying layers. Total loss is calculated with the aggregate of the data loss and residual loss.}
        \label{tab:quad_30neurons_total}
    \end{table}

In Table \ref{tab:quad_30neurons_total}, we see that 5 layers, once again, is the optimal configuration with a lower average total loss at $6.547 * 10^{-3}$ and a reasonably low standard deviation at $4.308*10^{-3}$. However, a model with 6 layers also seems ideal, as the different between 5 and 6 layers is minimal and the standard deviation at 6 layers is smaller than the standard deviation at 5 layers. Ultimately, we decided to choose 5 layers due to another variable\textemdash Time.

Considering the time between 5 and 6 layers, which were $39.779$ seconds and $46.648$ seconds, respectively. Evidently 5 layers is the most computing and time efficient and due to the minimal difference between the 2 configurations with respect to total loss and standard deviation we decided to continue with 5 layers as our optimal model.

In Table \ref{tab:quad_4layers_interval} we examined the difference between the ground truth MSE inside the interval $[0, 1]$ and $[-1, 2]$ to determine how well our model can understand trends in data both inside and outside the training interval. We used the ground truth MSE to compare our model to the trends of the data instead of the data itself. As explained earlier in Section \ref{subsec:optimallinear}, we generally want to use the interval, $[-1, 2]$ in our tests, as our goal is to capture trends outside the interval where the model was trained.

    \begin{table}[htbp]
        \centering
        \begin{tabular}{c|c|c|c}
        Layers & Neurons & $[-1, 2]$ Ground Truth MSE & $[0, 1]$ Ground Truth MSE\\
         \hline
        4 & 5 & $1.468 *10^{-2}$ & $6.425*10^{-3}$\\
        \hline
        4 & 10 & $7.900*10^{-3}$ & $5.875*10^{-3}$\\
        \hline
        4 & 20 & $7.611*10^{-3}$ & $5.963*10^{-3}$ \\
        \hline
        4 & 30 & $\boldsymbol{7.564*10^{-3}}$ & $\boldsymbol{5.790*10^{-3}}$ \\
        \hline
        4 & 40 & $7.646*10^{-3}$ & $5.912*10^{-3}$\\
        \hline
        2 & 30 & $1.251*10^{1}$ & $1.107*10^{-2}$\\
        \hline
        3 & 30 & $1.008*10^{-1}$ & $9.543*10^{-3}$\\
        \hline
        4 & 30 & $6.616*10^{-2}$ & $6.221*10^{-3}$ \\
        \hline
        5 & 30 & $\boldsymbol{3.157*10^{-2}}$ & $\boldsymbol{6.129*10^{-3}}$ \\
        \hline
        6 & 30 & $5.132*10^{-2}$ & $6.319*10^{-3}$\\
        \end{tabular}
        \caption{Ground truth MSE for one iteration calculated in two ways: throughout the interval $[-1, 2]$ and exclusively within the training interval $[0, 1]$. The first method was used for determining optimal layers and neurons. }
        \label{tab:quad_4layers_interval}
    \end{table}

As shown in Table \ref{tab:quad_4layers_interval}, we can measure ground truth MSE in two ways and should be aware of how our choice impacts the decided layer and neuron configuration. In this run, both the optimal number of neurons and layers stayed the same on both intervals with 30 neurons for both intervals $[-1, 2]$ and $[0, 1]$. When 30 neurons was held constant, 5 layers was optimal according to intervals $[-1, 2]$ and $[0, 1]$. From this information, we understood that our criterion for deciding layers and neurons can impact the final decision. Despite this limitation, the MSE values are similar and performance may not vary significantly, however, 5 layers and 30 neurons consistently prove to be the best model.

We also find it important to note that just like the Linear PINN there is a discrepancy in average ground truth MSE for 5 layers and 30 neurons between Table \ref{tab:quad_4layers} and Table \ref{tab:quad_30neurons}. The reason for the difference is that each run was random, and the averages were calculated over multiple iterations; the variation is due to instances of uncertainty between two values after 5 iterations. When we had doubt, we ran more iterations to gain additional context. The standard deviation in those tables explain that the data was separately calculated.

Our final optimal network architecture for the linear PINN was 5 layers and 30 neurons.

\subsection{Details on Heat Equation PINN Architecture}\label{subsec:optimalheat_appendix}
Here we provide a closer look at the method outlined in Section \ref{subsec:nn} and more specifically, Section \ref{subsubsec:heat_optimalln}. We also analyze alternative ways to find a satisfactory layer and neuron configuration. In making our decision, we ran the different models over 2500 epochs rather than 5000 epochs as we had done for the linear and quadratic models; this choice was influenced by the runtime of these different models. 

According to the mentioned method, we begin with 5 different models that all have 4 layers and a varying number of neurons per layer. As shown in Table \ref{tab:heat_4l_gt}, the optimal number of neurons is 80 when we consider the average ground truth MSE. This makes sense to us because the data is more complex and may require additional neurons to interpret patterns in it.

\begin{table}[htbp]
    \centering
    \begin{tabular}{c|c|c|c}
        Layers & Neurons & Avg. Ground Truth MSE & Standard Deviation\\
         \hline
        4 & 10 & $3.702*10^{-7}$ & $1.339*10^{-7}$\\
        \hline
        4 & 20 & $1.654*10^{-7}$ & $1.249*10^{-7}$\\
        \hline
        4 & 40 & $3.810*10^{-8}$ & $1.683*10^{-8}$ \\
        \hline
        4 & 60 & $9.950*10^{-9}$ & $5.993*10^{-9}$ \\
        \hline
        4 & 80 & 
        $\boldsymbol{7.172*10^{-9}}$ & $3.233*10^{-9}$\\
    \end{tabular}
    \caption{Average MSE calculated with ground truth function over $[0, 1]$ and its standard deviation. Error is calculated for models with 4 fixed layers and varying neurons. The data is averaged over 5 iterations.}
    \label{tab:heat_4l_gt}
\end{table}

However, in most situations, the ground truth function is not known. In those cases, we would rely on the total loss containing the residual to make our decision. Table \ref{tab:heat_4l_total} outlines this possibility. The table tells us that even if we decided the optimal number of neurons using the total loss, we would still choose 80 neurons. This consistency further strengthened our decision of 80 neurons. 

If there was a discrepancy, we would have still made a decision based on the average ground truth MSE because it is more reliable; the average ground truth MSE is evaluated on 300 data points across the interval, whereas the average total loss is evaluated on 10. 

\begin{table}[htbp]
    \centering
    \begin{tabular}{c|c|c|c}
        Layers & Neurons & Avg. Total Loss & Standard Deviation\\
         \hline
        4 & 10 & $5.216*10^{-5}$ & $1.010*10^{-5}$\\
        \hline
        4 & 20 & $1.638*10^{-5}$ & $7.464*10^{-6}$\\
        \hline
        4 & 40 & $4.867*10^{-6}$ & $1.691*10^{-6}$ \\
        \hline
        4 & 60 & $1.823*10^{-6}$ & $5.809*10^{-7}$ \\
        \hline
        4 & 80 & 
        $\boldsymbol{1.301*10^{-6}}$ & $4.762*10^{-7}$\\
    \end{tabular}
    \caption{Average total loss over $[0, 1]$ and its standard deviation. Error is calculated for models with 4 fixed layers and varying neurons. The data is averaged over 5 iterations.}
    \label{tab:heat_4l_total}
\end{table}

Then, we varied the number of layers while fixing 80 neurons. Table \ref{tab:heat_80n_gt} displays those ground truth MSE results, which communicate that 5 layers is the optimal number of layers for 80 neurons. 

\begin{table}[htbp]
    \centering
    \begin{tabular}{c|c|c|c}
        Layers & Neurons & Avg. Ground Truth MSE & Standard Deviation\\
         \hline
        2 & 80 & $2.151*10^{-5}$ & $5.889*10^{-6}$\\
        \hline
        3 & 80 & $2.164*10^{-7}$ & $4.125*10^{-8}$\\
        \hline
        4 & 80 & $7.765*10^{-8}$ & $4.398*10^{-8}$ \\
        \hline
        5 & 80 & $\boldsymbol{3.139*10^{-8}}$ & $9.159*10^{-9}$ \\
        \hline
        6 & 80 & 
        $3.337*10^{-8}$ & $1.070*10^{-8}$\\
    \end{tabular}
    \caption{Average MSE calculated with ground truth function over $[0, 1]$ and its standard deviation. Error is calculated for models with 80 fixed neurons and varying layers. The data is averaged over 5 iterations.}
    \label{tab:heat_80n_gt}
\end{table}

Again, we compared these results to those using the final total loss from training as our criterion. Table \ref{tab:heat_80n_total}, similar to Table \ref{tab:heat_4l_total}, shows us consistency; even when considering the average total loss, 5 layers is still optimal when there are 80 neurons per layer.

\begin{table}[htbp]
    \centering
    \begin{tabular}{c|c|c|c}
        Layers & Neurons & Avg. Total Loss & Standard Deviation\\
         \hline
        2 & 80 & $1.762*10^{-3}$ & $5.500\times10^{-4}$\\
        \hline
        3 & 80 & $2.616*10^{-5}$ & $8.776*10^{-6}$\\
        \hline
        4 & 80 & $8.254*10^{-6}$ & $2.284*10^{-6}$ \\
        \hline
        5 & 80 & $\boldsymbol{5.283*10^{-6}}$ & $8.698*10^{-7}$ \\
        \hline
        6 & 80 & 
        $5.406*10^{-6}$ & $1.479*10^{-6}$\\
    \end{tabular}
    \caption{Average total loss over $[0, 1]$ and its standard deviation. Error is calculated for models with 80 fixed neurons and varying layers. The data is averaged over 5 iterations.}
    \label{tab:heat_80n_total}
\end{table}

In the case of the heat equation, the model only functions on the $[0, 1]$ range. This limitation is due to the way we enforced boundary and initial conditions at the endpoints. Because of this, we do not have an alternate way to calculate ground truth MSE as we did for the linear and quadratic models.

We also find it important to note that for the combination of 4 layers and 80 neurons, there are discrepancies between the average ground truth MSEs in Table \ref{tab:heat_4l_gt} and Table \ref{tab:heat_80n_gt} as well as between the average total losses in Table \ref{tab:heat_4l_total} and \ref{tab:heat_80n_total}. This difference is because each run of the model is random, and the averages are calculated over only five iterations in the interest of computational efficiency. Thus, the averages may not converge to the same number. For this reason, we also reported the standard deviations to add context to the data.

\subsection{Details on FDM and PINN Comparisons}\label{subsec:fdm_pinn}

Here we provide a closer look at our runs of various dx, dt, D, and T values with the FDM and PINN.

\begin{table}[htbp]
    \centering
    \begin{tabular}{c|c|c}
        D & PINN Data Point MSE & FDM MSE \\
        \hline
        0.01 & 4.293E-10 & 1.457E-08 \\
        \hline
        0.05 & 2.010E-09 & 5.169E-06 \\
        \hline
        0.09 & 7.310E-09 & 4.764E-05 \\
    \end{tabular}
    \caption{Comparison of DP MSE, Total Loss, and FDM MSE for different D Values when dx=dt=0.1 and T=1}
    \label{table:dp_fdm_comparison}
\end{table}

Here, in Table \ref{table:dp_fdm_comparison}, we first look at this combination and see that the PINN performs well, consistently yielding mean squared error calculated at the data points well below error from the FDM. At this density, if the PINN and FDM both followed the stability conditions of the CFL condition, we would see the errors diverging rapidly after D=0.05, but we can see that the PINN still performs well at D=0.09.

\begin{table}[htbp]
    \centering
    \begin{tabular}{c|c|c}
        D & PINN Data Point MSE & FDM MSE \\
        \hline
        0.01 & 6.268E-08 & 3.996E-07 \\
        \hline
        0.05 & 5.330E-07 & 6.329E-06 \\
        \hline
        0.09 & 3.745E-05 & 1.084E+44 \\
    \end{tabular}
    \caption{Comparison of DP MSE, Total Loss, and FDM MSE for different D Values when dx=dx=0.1 and T=10}
    \label{table:dp_fdm_comparison2}
\end{table}

Here, in Table \ref{table:dp_fdm_comparison2}, we then observe the same dx, dt, and D values, but change our overall time to T=10. We can much better see the FDM blowing up at D=0.09 at a later time point. Once again, the PINN performs well in comparison.

\begin{table}[htbp]
    \centering
    \begin{tabular}{c|c|c}
        D & PINN Data Point MSE & FDM MSE \\
        \hline
        0.01 & 4.849E-03 & 1.268E-07 \\
        \hline
        0.05 & 1.605E-03 & 6.404E-07 \\
        \hline
        0.09 & 7.185E-01 & Limits too large \\
    \end{tabular}
    \caption{Comparison of DP MSE, Total Loss, and FDM MSE for different D Values when dx=dt=0.1 and T=100}
    \label{table:dp_fdm_comparison3}
\end{table}

In Table \ref{table:dp_fdm_comparison3}, we can even more clearly observe the FDM diverging when our maximum time is at T=100.

Then, we moved on to higher densities. First we look at the FDM and PINN when dx=dt=0.05.

\begin{table}[htbp]
    \centering
    \begin{tabular}{c|c|c}
        D & PINN Data Point MSE & FDM MSE \\
        \hline
        0.005 & 4.015E-10 & 2.493E-10 \\
        \hline
        0.025 & 2.428E-09 & 1.167E-07 \\
        \hline
        0.045 & 3.101E-08 & 1.382E-06 \\
    \end{tabular}
    \caption{Comparison of DP MSE, Total Loss, and FDM MSE for different D Values when dx=dt=0.05 and T=1}
    \label{table:dp_fdm_comparison4}
\end{table}

Here, in Table \ref{table:dp_fdm_comparison4}, we see that both models seem to perform well when our model ends at T=1.

\begin{table}[htbp]
    \centering
    \begin{tabular}{c|c|c}
        D & PINN Data Point MSE & FDM MSE\\
        \hline
        0.005 & 5.134E-08 & 1.277E-08 \\
        \hline
        0.025 & 5.613E-08 & 7.127E-07 \\
        \hline
        0.045 & 1.558E-07 & 3.520E+128 \\
    \end{tabular}
    \caption{Comparison of DP MSE, Total Loss, and FDM MSE for different D Values when dx=dt=0.05 and T=10}
    \label{table:dp_fdm_comparison5}
\end{table}

\pagebreak
In \ref{table:dp_fdm_comparison5}, we see that with a D value of 0.045, the FDM clearly blows up when we let the model run over a higher t value of T=10. The PINN, however, still works well.

\begin{table}[htbp]
    \centering
    \begin{tabular}{c|c|c}
        D & PINN Data Point MSE & FDM MSE \\
        \hline
        0.005 & 1.405E-03 & 1.633E-08 \\
        \hline
        0.025 & 2.951E-04 & 8.213E-08 \\
        \hline
        0.045 & 5.050E-04 & Limits too large \\
    \end{tabular}
    \caption{Comparison of DP MSE, Total Loss, and FDM MSE for different D Values when dx=dt=0.05 and T=100}
    \label{table:dp_fdm_comparison6}
\end{table}

In Table \ref{table:dp_fdm_comparison6}, we see once again that our MSe and total loss for our PINN and FDM do increase, but only the FDM dramatically diverges when the model is allowed to run until T=100.

Finally, we looked at changing D values but with an even higher density, with dx=dt=0.0025

\begin{table}[htbp]
    \centering
    \begin{tabular}{c|c|c}
        D & PINN Data Point MSE & FDM MSE\\
        \hline
        0.0025 & 2.349E-10 & 4.085E-12 \\
        \hline
        0.0125 & 1.085E-09 & 2.206E-09 \\
        \hline
        0.0225 & 6.872E-10 & 2.944E-03 \\
    \end{tabular}
    \caption{Comparison of DP MSE, Total Loss, and FDM MSE for different D Values when dx=dt=0.025 and T=1}
    \label{table:dp_fdm_comparison7}
\end{table}

In Table \ref{table:dp_fdm_comparison7}, we see again that the FDM has a much higher error than the PINN but is not quite exploding at a small time of T=1.

\begin{table}[htbp]
    \centering
    \begin{tabular}{c|c|c}
        D & PINN Data Point MSE & FDM MSE\\
        \hline
        0.0025 & 2.610E-08 & 2.909E-10 \\
        \hline
        0.0125 & 2.580E-06 & 4.687E-08 \\
        \hline
        0.0225 & 1.249E-07 & 3.242E+294 \\
    \end{tabular}
    \caption{Comparison of DP MSE, Total Loss, and FDM MSE for different D Values when dx=dt=0.025 and T=10}
    \label{table:dp_fdm_comparison8}
\end{table}

\pagebreak
In Table \ref{table:dp_fdm_comparison8}, the same pattern follows. Only the FDM explodes rapidly at high time values.

\begin{table}[htbp]
    \centering
    \begin{tabular}{c|c|c}
        D & PINN Data Point MSE & FDM MSE \\
        \hline
        0.0025 & 2.367E-04 & 1.819E-09 \\
        \hline
        0.0125 & 1.976E-04 & 1.046E-08 \\
        \hline
        0.0225 & 9.380E-05 & Limits too large \\
    \end{tabular}
    \caption{Comparison of DP MSE, Total Loss, and FDM MSE for different D Values when dx=dt=0.025 and T=100}
    \label{table:dp_fdm_comparison9}
\end{table}

In Table \ref{table:dp_fdm_comparison9}, we see that the limits of the model are exceeded from time 10 to 100. The PINN error gets larger but does not explode like the FDM.

We conclude that the PINN does not fit the same conditions as the FDM and other traditional methods of modeling the heat equation. We suspect that PINNs are much more tolerable of high densities and D values. We will try to find the range and breaking point of the PINN with the heat equation.

\printbibliography

\end{document}